\documentclass[acmlarge,nonacm,review=false,timestamp=false]{acmart}
\AtBeginDocument{%
  }

\usepackage{booktabs}
\usepackage{tabularx}

\begin{document}

\title{Towards Pervasive Distributed Agentic Generative AI - A State of The Art}

\author{Gianni Molinari}
\email{gianni.molinari@unito.it}
\orcid{0009-0000-5660-3407}
\affiliation{%
  \institution{University of Turin}
  \city{Turin}
  \country{Italy}
}

\author{Fabio Ciravegna}
\affiliation{%
  \institution{University of Turin}
  \city{Turin}
  \country{Italy}}
\email{fabio.ciravegna@unito.it}
\orcid{0000-0001-5817-4810}

\begin{abstract}
  The rapid advancement of intelligent agents and Large Language Models (LLMs) is reshaping the pervasive computing field. Their ability to perceive, reason, and act through natural language understanding enables autonomous problem-solving in complex pervasive environments, including the management of heterogeneous sensors, devices, and data. This survey outlines the architectural components of LLM agents (profiling, memory, planning, and action) and examines their deployment and evaluation across various scenarios. Then it reviews computational advancements (cloud to edge) in pervasive computing and how AI is moving in this field. It highlights state-of-the-art agent deployment strategies and applications, including local and distributed execution on resource-constrained devices. This survey identifies key challenges of these agents in pervasive computing such as architectural, energetic and privacy limitations. It finally proposes what we called "\textbf{Agent as a Tool}", a conceptual framework for pervasive agentic AI, emphasizing context awareness, modularity, security, efficiency and effectiveness.
\end{abstract}




\maketitle

\section{Introduction}
Agentic generative AI systems integrate generative foundation models (e.g., large language and multi-modal models) as the core reasoning engines inside autonomous agents. Designed to perceive, reason, and act within dynamic environments, these agents are pivotal for the successful decomposition and execution of complex, high-level objectives in areas previously inaccessible to static models. The advent of Large Language Models (LLMs) has recently revolutionised this domain. Trained on vast web datasets, LLMs demonstrate an impressive understanding of human language and can generate remarkably human-like and accurate responses. Integrating these capabilities has led to the development of LLM-based agents, where LLMs serve as the core cognitive engine, combined with perceptual, reasoning, and action mechanisms. This synergy has introduced a new paradigm of more intelligent and versatile agents applicable across diverse domains.
Agents can have various capabilities. Some are purely reactive, responding to explicit user prompts~\cite{franciscatto2025cbr,zhang2024webpilotversatileautonomousmultiagent}, while others show proactive functionalities by autonomously initiating tasks based on their understanding of the environment~\cite{zhang2024proagent}. Their applications are widespread. Embodied agents interact with the physical world via sensors and actuators~\cite{durante2024agent}, whereas software agents operate in digital environments, performing tasks like information retrieval or software and research development~\cite{zhang2024aflow, zhang2024agentic}.

Also the field of pervasive computing, focused on integrating computational and communication environments into daily human life, has seen substantial progress. This domain includes a wide spectrum of systems, ranging from high-performance computing (HPC) to resource-constrained IoT devices, with applications spanning smart homes, factories, and hospitals.
Systems deployed in this pervasive context must be able to execute complex, high-level tasks. Such execution requires extraction and synthesis of information from diverse structures (tabular, document-based) and modalities (textual, images), coupled with seamless interaction with the surrounding environment.
Consequently, LLM-based autonomous agents represent an architectural solution for this necessity. By design, these agents possess the fundamental capabilities to perceive from various information sources, reason the actions to take, and interact with surrounding systems using external actuators (often referred to as tools). This unique combination of capabilities give to agentic systems an enormous potential to operate effectively and autonomously within pervasive computing domains.

However, deploying agents in these environments presents significant computational, and security challenge that necessitate custom solutions. Researchers have proposed various solutions to solve these challenges.
This paper analyses strategies for integrating agents into pervasive computing, examines existing applications, identifies persistent challenges, and explores ongoing research efforts to address them.
Specifically, the paper is structured as follows:
\begin{enumerate}
    \item First, an introduction to LLM-based agent systems cover their architecture, performance evaluation metrics for accuracy and robustness, and diverse applications.
    \item Next, the paper will introduce pervasive computing, detailing recent advancements, its underlying infrastructure, and the integration of artificial intelligence, culminating in agent adoption.
    \item Then, it will explore agent implementation in pervasive environments, focusing on architectural strategies for effective deployment tailored to available infrastructure, supported by application examples.
    \item A detailed discussion will then analyse unresolved challenges in the field and current research directions aimed at their resolution.
    \item Finally, the paper will propose a future research vision for LLM-Agents in pervasive computing, culminating in the "\textbf{Agent as a Tool}" concept.
\end{enumerate}

\section{Agent LLM Architecture}
\label{sec:agents}
\begin{figure}[htbp]
    \centering
    \includegraphics[width=0.9\textwidth]{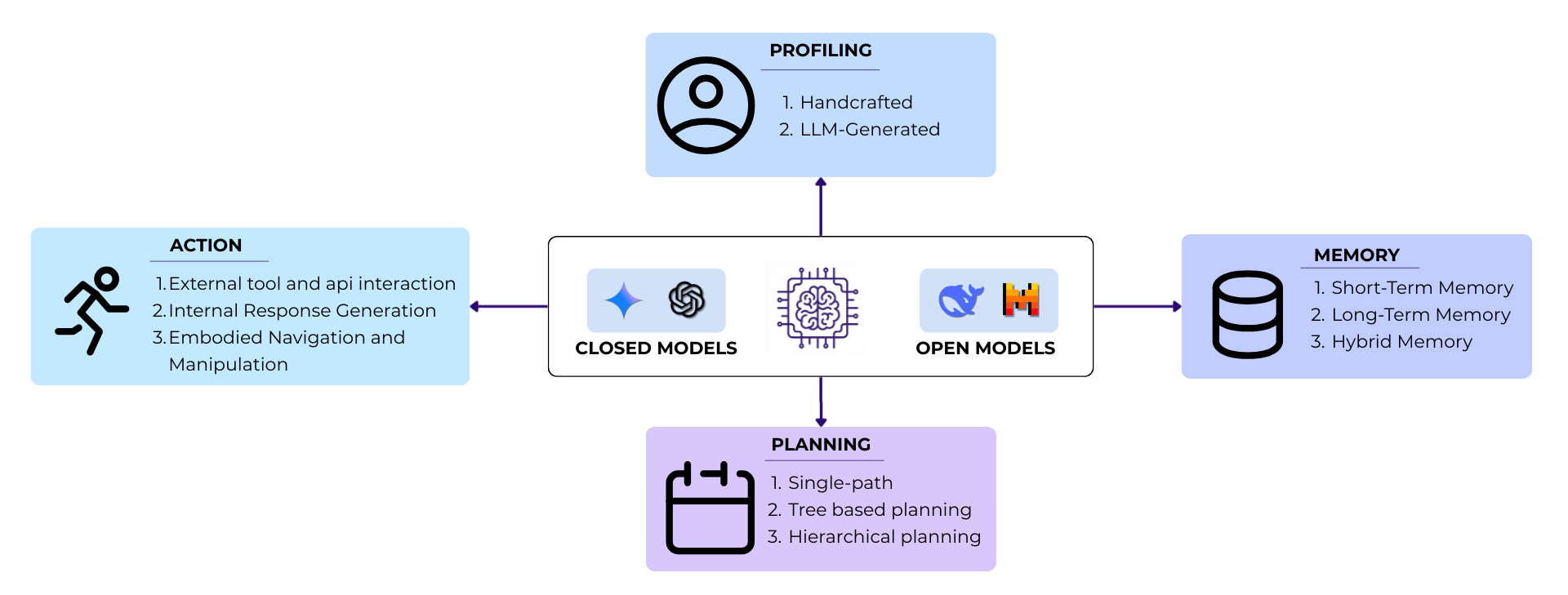}
    \caption{The LLM Agent architecture}
    \label{fig:llmagent}
\end{figure}

In the context of LLM-based agents, a natural language processing pipeline is employed to enable the agent to perceive, reason, and interact with its environment to accomplish specific goals. Perception, the initial stage in this process, involves gathering information about the current state of the environment~\cite{wang2024survey}. This often occurs through natural language inputs, where users provide instructions or queries~\cite{franciscatto2025cbr}, or structured text, including HTML or a programming language formats~\cite{hu2024automated, zhang2024webpilotversatileautonomousmultiagent}. In addition, multimodal LLMs (MLLM), process visual and auditory data, resulting in enhanced environmental comprehension and information use~\cite{durante2024agent}. Also, to enhance environmental interaction, respect constraints, and achieve goals, LLM-based agents require a robust understanding of implied textual meanings~\cite{li2024lasp}.
Following perception, the agent leverages the Large Language Model's (LLM) natural language processing capabilities for reasoning and planning. This entails processing the perceived information, interpreting user intent, and analyzing possible courses of action~\cite{parmar2025plangen, wang2024survey}. A key aspect of this stage is the formulation of strategic plans to achieve desired outcomes. Complex tasks are often decomposed into manageable subtasks to facilitate execution~\cite{li2024lasp, parmar2025plangen} such as for robots movements management~\cite{yang2025magma}, automatic world exploration~\cite{wang2023voyager} or code debugging~\cite{wang2024openhands}. The final stage involves in interacting directly with the environment following the previous reasoning steps~\cite{wang2024survey}. Actions may include generating natural language responses to the user, providing explanations or completing tasks through dialogue~\cite{zhang2024agentic, xi2025rise}, invoke external tools or APIs~\cite{durante2024agent, schick2023toolformer} or produce executable code~\cite{zhang2024aflow}. 
To accomplish all these tasks modern agent architectures typically consist of multiple key modules that work together to create intelligent behavior. A typical LLM-agent architecture has four key modules: a profiling module for defining the agent's role, a memory module for storing and retrieving past experiences, a planning module for formulating future actions, and an action module for translating decisions into outputs~\cite{zhang2024agentic}. 
Also, evaluating this architecture is essential to determine the effectiveness of its modules, especially within the intended application domain.
The following sections will explore the LLM architecture, the specific functionalities and interactions of the agent modules, and the evaluation strategies and application domains of these intelligent LLM agents.

\subsection{Large Language Models}
\begin{figure}[htbp]
    \centering
    \includegraphics[width=0.3\textwidth]{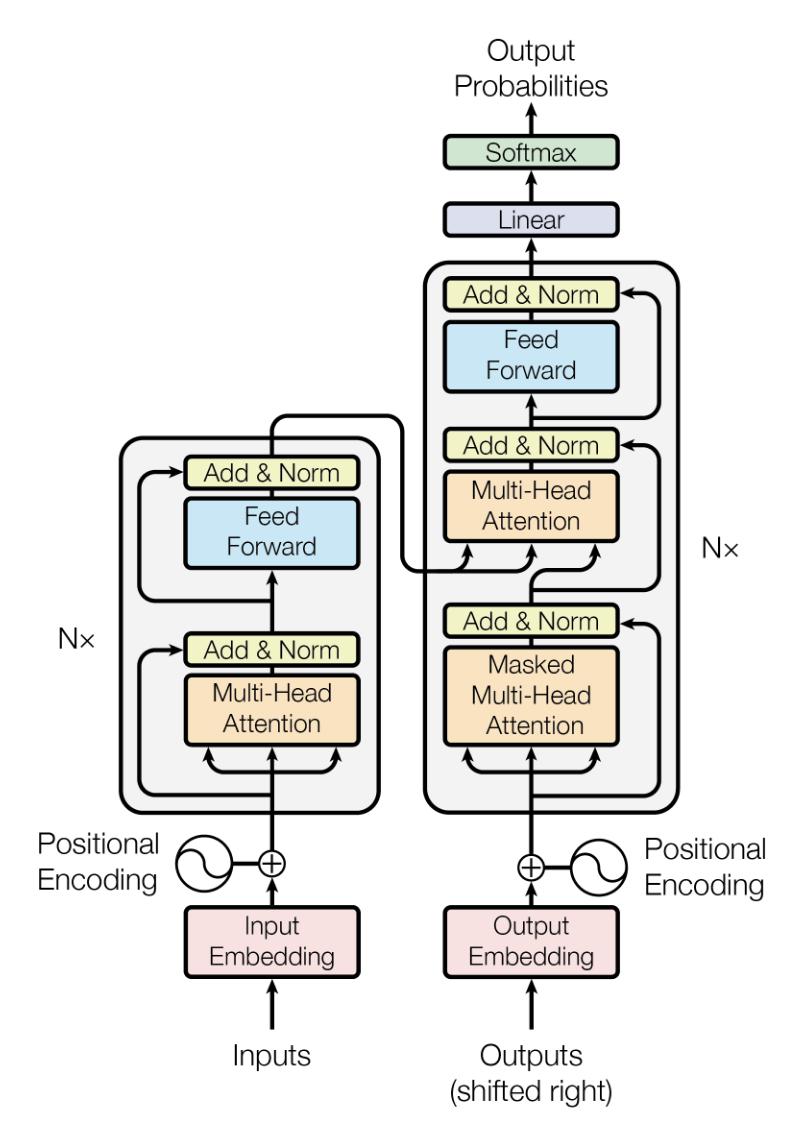}
    \caption{The transformer architecture, from~\cite{vaswani2017attention}}
    \label{fig:transformer}
\end{figure}

The core cognitive component, or "brain", of intelligent agents are Large Language Models (LLMs). Their fundamental operation rely on "next token prediction". This means that given an input sequence of words (tokens), transformed into numerical embeddings, LLMs predict the subsequent token in the sequence, iteratively generating the output.
The success of LLMs in this task is largely attributed to their foundation in the Transformer architecture, illustrated in Figure~\ref{fig:transformer}. Its design captures long-range dependencies within the text and effectively mimics the complex structures of human language~\cite{vaswani2017attention}. A typical LLM consists of stacked Transformer layers, each incorporating a multi-head self-attention mechanism and a position-wise fully connected feed-forward network.
The multi-head self-attention mechanism allows the model to weigh the importance of different parts of the input sequence when processing each word. It processes the encoded input as a set of key-value pairs (K, V), where both keys and values have a dimension equal to the input sequence length (n). The Transformer employs scaled dot-product attention to calculate how much attention each value should receive based on a query (Q):
$$Attention(Q,K,V)=softmax(\frac{QK^T}{\sqrt{n}}V)$$
Instead of performing this attention calculation just once, the multi-head mechanism does it in parallel multiple times. The resulting independent attention outputs are then combined (concatenated) and transformed linearly to produce the desired output dimensions.
The Transformer architecture is fundamentally composed of two main components: an encoder and a decoder. The encoder's role is to create an attention-based representation of the input, enabling it to "attend" to relevant information. Each encoder layer includes a multi-head self-attention sub-layer and a simple position-wise feed-forward network, with each sub-layer benefiting from a residual connection and layer normalization.
The decoder's function is to generate the output sequence based on the encoded representation. Each decoder layer includes two multi-head attention sub-layers (one for self-attention and one for attention over the encoder's output) and a feed-forward network, again with residual connections and layer normalization. Finally, a a linear layer and a softmax are applied to the decoder's output, producing a probability distribution over the vocabulary from which the next word is sampled.

LLMs commonly employ this Transformer architecture, scaling it to billions or even trillions of parameters and incorporating various optimization strategies. For instance, decoder-only architectures like GPT-3 and LLaMA predict each next token based on the preceding ones. Conversely, encoder-only architectures such as BERT and RoBERTa prioritise understanding the input text to generate task-specific outputs like labels or token predictions~\cite{matarazzo2025survey}.
In the agentic field, the choice of architecture depends on the required task. Typically, decoder-only architectures are most commonly used when we want an agent to perceive, reason, and interact with the environment.
Furthermore, Multimodal Language Models (MLLMs) are increasingly gaining traction for agents interacting in multimodal environments (text, images, video, audio). This is because they allow the application of Transformer architectures across diverse modalities like vision and audio. 
To enable such multimodal capabilities, MLLMs extend the Transformer architecture by embedding non-text inputs (e.g., image patches or audio tokens) into the same vector space as textual tokens. These embeddings are then processed jointly within shared Transformer layers or through modality-specific experts, such as in Multiway Transformer architectures. Deep modality fusion is achieved via cross-attention or unified encoders, enabling fine-grained interactions between visual and textual features~\cite{liang2024survey}.

\subsection{Agent Modules}
LLM-based autonomous agents are designed to effectively perform diverse tasks by leveraging the capabilities of Large Language Models (LLMs). To achieve this, their architecture typically incorporates 4 key modules (Figure \ref{fig:llmagent}): profiling module, memory module, planning module, and action module. 

\subsubsection{Profiling}
\begin{figure}[htbp]
    \centering
    \includegraphics[width=0.6\textwidth]{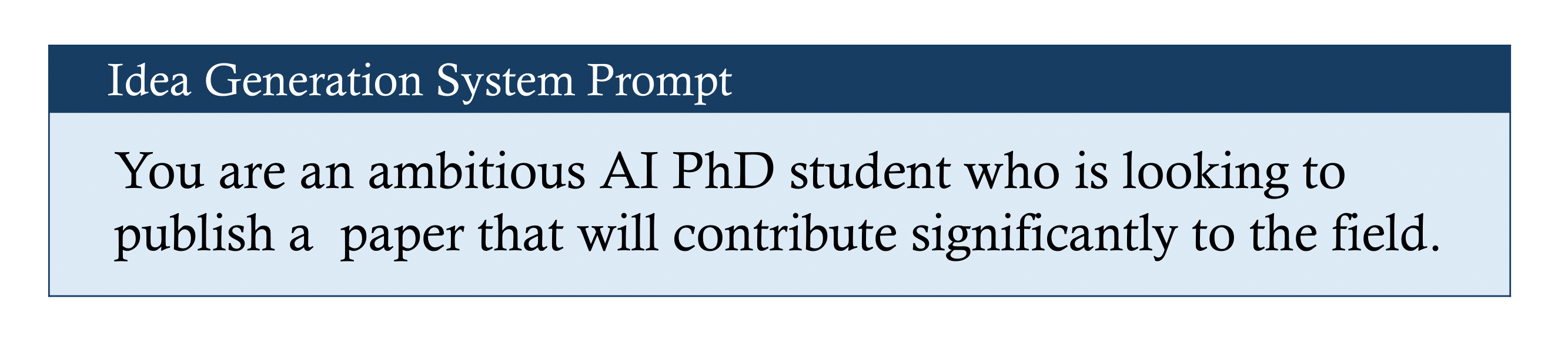}
    \caption{An example of profiling prompt used in~\cite{lu2024ai}}
    \label{fig:profile}
\end{figure}
The primary purpose of the Profiling Module is to define the role-specific identity of the agent, which significantly influences its behavior and interactions~\cite{wang2024survey, zhang2024agentic}. This module is crucial for enabling agents to generate more contextually relevant responses from the LLM across different modalities. These profiles are typically added into the model context window. By including role descriptions in the prompt, the profiling module guides the LLM's reasoning style, response formulation, and interaction strategy~\cite{zhang2024agentic}. For instance, in~\cite{lu2024ai} the agent is profiled to work as phD student that wishes to publish some papers (Figure~\ref{fig:profile}). The profiling module serves as the foundation for agent design, giving a significant influence on the entire agent's modules ecosystem~\cite{wang2024survey}. The chosen profile can impact how the agent remembers information (memory module), how it formulates strategies to achieve its goals (planning module) that consequently will impact also the future actions. For example, an agent profiled as '\textit{an expert machine-learning researcher}'~\cite{hu2024automated} might prioritise certain experiments over others or selectively memorise results more significant to its research. Similarly, an agent profiled as '\textit{an advanced AI system serving as an impartial judge for intelligent code generation outputs}'~\cite{zhuge2024agent} would focus more intently on code details, errors, and bugs than on other capabilities. 
Agent profiles are commonly implemented by handcrafting specific prompts that describe the desired role and behavior~\cite{wang2024survey}. However, other strategies can be used, including LLM-generation of profiles and dataset alignment to reflect real-world characteristics~\cite{wang2024survey}. Combining these strategies can yield additional benefits. Moreover agent profiles can include multiple dimensions, including basic attributes, behavioral patterns, and social information~\cite{zhang2024agentic}.

\subsubsection{Memory} 
Large Language Models lack persistent memory across test time interactions, requiring prior exchanges to be explicitly included within the context window (Figure~\ref{fig:memory_schema}). This limitation underscores the need for effective memory management strategies in LLM-based agents, particularly for prioritizing relevant information during storage and retrieval. As the agent's interaction history expands, exceeding the fixed context window's capacity, summarization or information filtering becomes unavoidable.
Consequently, autonomous LLM-based agents rely on a memory module to retain, recall, and reflect upon past observations, thoughts, feedback, and actions. This enables them to better understand the current context, anticipate future needs, correct their action trajectory, adjust their behavior, and maintain consistent and coherent behavior over time~\cite{wen2024autodroidv1, xi2025rise, wang2024survey}.
This module often incorporate different structures to manage information across varying timescales:
\begin{itemize}
    \item \textbf{Short-Term Memory:} This type of memory maintains contextually relevant information about recent perceptions, reasoning and actions (the trajectory history) within the LLM context window~\cite{zhang2024agentic,shinn2023reflexion}. It allows for real-time adaptation during an interactive session. Examples include the internal states maintained by a conversational agent or the intermediate steps during a web research~\cite{zhang2024webpilotversatileautonomousmultiagent}. However, the limited context window of LLMs poses a challenge for relying solely on short-term memory.
    \item  \textbf{Long-Term Memory:} In contrast to the real-time encounter observations, which provide short-term memory, the overall encounter history represents long-term memory~\cite{zhang2024survey}. Long-term memory can provide stable knowledge to complement the flexibility of short-term memory. This module is important because stores both successful experiences, demonstrating correct goal achievement strategies, and unsuccessful experiences, highlighting incorrect paths to avoid. For instance in~\cite{wu2024beyond}, successful paths are stored as 'Thought Cards' and retrieved upon encountering a new question with a similar topic or in~\cite{zhang2024proagent} the agent store historical events that will be used to predict potential tasks.
    \item \textbf{Hybrid Memory:} Many agents use a combination of short-term and long-term memory to leverage the benefits of both~\cite{wang2024survey}. Short-term memory handles immediate context, while long-term memory provides access to a broader history and consolidated knowledge.
\end{itemize}

These structures can be stored in various formats, each with its own advantages. 
One approach is to keep everything in natural language form. This is especially useful for conversational agents, where it's important to be transparent about the agent's history. For instance, the method in~\cite{shinn2023reflexion} stores past trials as verbal descriptions, helping the model to understand and correct previous attempts. Similarly, in~\cite{wang2023voyager} descriptions of skills within the Minecraft game are stored directly as natural language within the agent's memory.
Then, there's the idea of Embedding Memory. Here, past experiences are transformed into high-dimensional vectors~\cite{wang2024survey}. This allows for really fast reflection and helps the agent generalise across different situations. It's especially good for searching through knowledge quickly, based on semantic similarity~\cite{zhang2024agentic}.
Finally databases can be used to store memories. This gives us considerable power to manipulate the data through structured queries. So agents can be used to understand and execute SQL queries in natural language, which lets it interact with the database really effectively~\cite{shen2024demonstration}.

Three core operations that can be applied to the memory. First writing, which involves persisting relevant information from past interactions into the memory. Key considerations during memory writing include handling duplicated information and managing memory overflow~\cite{chen2024llm}. Various techniques are used such as summarizing similar information, using a FIFO buffer to overwrite old entries~\cite{wang2024survey}, truncating historical data that exceeds processing capacity~\cite{xi2025rise}, and implementing time limits to discard redundant data and prevent excessive memory consumption~\cite{zhong2024casit}.
Next, memory reading is a crucial operation that aims to obtain relevant knowledge from the agent's memory to adapt its behaviours and inform its next actions~\cite{zhang2024agentic}. The retrieval of information often considers factors like recency, relevance, and importance of the memories. These factors commonly include the recency of the memory, its relevance to the current situation or query, and its perceived importance~\cite{xi2025rise, wang2024survey}.
Finally, memory reflection allows the agent to analyse, refine, and optimise the accumulated memories~\cite{zhang2024agentic}. It emulates human cognitive processes of summarizing and inferring more abstract and high-level insights from past experiences. Reflection can occur hierarchically, generating novel based on existing insights~\cite{shinn2023reflexion}.
As mentioned in the previous section, the memory module is not isolated. It is influenced by the agent's profile, which can determine what types of data are prioritised for storage and retrieval. Furthermore, the memory module plays a critical role in the planning process, providing the necessary context and historical data for the agent to formulate effective future actions. The planning module might retrieve information from memory to inform its reasoning and decision-making.

\subsubsection{Planning}
\label{sec:planning}
The planning module is essential for autonomous agents, enabling them to formulate future action sequences to achieve defined goals~\cite{wang2024survey}. It allows agents to move in goal-directed, multi-step problem-solving approach. The planning module combines environmental feedback, current state, desired outcomes, memory, and profile information to construct a trajectory of actions that effects a transition from the agent's current state to a target goal state.
Various methodologies have emerged for incorporating LLMs into the planning process, each representing a different approach to leveraging their capabilities: 
\begin{itemize}
    \item \textbf{LLM-as-Planner} directly employs the inherent reasoning abilities of LLMs to generate plans from natural language instructions~\cite{li2024lasp}. This paradigm emphasizes the autonomous planning capacity of LLMs, relying on their ability to interpret and translate natural language directives into actionable plans.
    \item \textbf{LLM-as-Facilitator} uses LLMs to augment existing planning algorithms, such as classic symbolic planners. In this context, LLMs may serve as translators, converting natural language problem descriptions into formal planning languages, such as the Planning Domain Definition Language (PDDL), which are then processed by external planners~\cite{li2024embodied, wang2024survey}.
    \item \textbf{Multi-Agent Planning} involves the coordination of plans and actions among multiple agents to achieve a shared objective~\cite{han2024llm}. This necessitates the implementation of mechanisms for inter-agent communication, negotiation, and belief revision. Collaborative planning approaches are exemplified by frameworks such as PlanGEN~\cite{parmar2025plangen}, which employs specialised LLM agents for constraint checking, verification, and selection, and Master~\cite{gan2025master}, which proposes a hierarchical multi-agent framework that dynamically generates collaborating agents, validates reasoning, and adjusts confidence-based scoring using Monte Carlo Tree Search (MCTS) to enhance accuracy and efficiency.
\end{itemize}
One key aspect of the planning module is task decomposition, where complex tasks are broken down into smaller, more tractable sub-problems, facilitating efficient execution. For example, PlanGEN~\cite{parmar2025plangen} is specifically engineered to augment LLMs' capacity to generate effective natural language plans through this decomposition process. Similarly, WebPilot's Planner module~\cite{zhang2024webpilotversatileautonomousmultiagent} initiates its operation by partitioning complex web tasks into manageable subtasks, thereby constructing a flexible, high-level plan that can adapt to the inherent uncertainties of web environments. Consequently, determining the optimal sequence of these subtasks or individual actions, becomes crucial for achieving the overarching goal. In more complex scenarios, the planning module may also necessitate the allocation of available resources across various planning components~\cite{li2024lasp}. This is exemplified by the TravelPlanner benchmark~\cite{xie2024travelplanner}, which evaluates agents' ability to generate travel itineraries from user queries, demanding the navigation of extensive online resources using specialised tools and adherence to user constraints. This benchmark assesses the agent's resource allocation across diverse tasks, including city searches, flight bookings, accommodation selection, and dining arrangements.
In the current literature the Planning Module employs diverse strategies in task decomposition:
\begin{itemize}
    \item \textbf{Single-Path Planning:} The agent follows one trajectory of thought and action at a time, without exploring alternative possibilities. Examples include Chain of Thought prompting, where the LLM reasons step-by-step along a linear path~\cite{wang2024survey}.
    \item \textbf{Tree-Based Planning:} The agent explores multiple potential thought trajectories, organizing them into a tree structure. This allows for backtracking and evaluating different options before committing to a plan. Techniques like Tree of Thoughts and methods using Monte Carlo Tree Search (MCTS) fall under this category~\cite{li2024lasp}.
    \item \textbf{Hierarchical Planning:} Involves planning at different levels of abstraction, with high-level plans broken down into more detailed, low-level actions or sub-goals~\cite{xi2025rise, zhang2024webpilotversatileautonomousmultiagent}. 

\end{itemize}

Effective integration of feedback within the planning module significantly enhances the ability of LLM-based agents to operate in complex and dynamic environments~\cite{han2024llm}. This capability allows agents to move beyond static, pre-defined plans and engage in a more adaptive and iterative problem-solving process. So we can categorise planning approaches in 2 groups:

\begin{itemize}
    \item \textbf{Planning without Feedback:} The agent formulates a plan upfront and executes it sequentially without adjusting based on intermediate outcomes~\cite{han2024llm}. This approach is appropriate for tasks of lower complexity, including conversational interaction.
    \item \textbf{Planning with Feedback:} The agent receives information from the environment, humans, or other models about the progress and outcomes of its actions~\cite{li2024lasp}. This feedback is then used to dynamically adjust and refine the plan. Environmental Feedback can include task completion signals or observations after taking an action~\cite{yao2023react}. Human feedback can provide guidance or corrections~\cite{franciscatto2025cbr}. Model-based feedback, such as self-reflection or critique from another AI model, can also be valuable~\cite{shinn2023reflexion}.
\end{itemize}

Finally, careful consideration of memory and profiling is vital for effective planning. The memory module provides the historical context, past experiences, and relevant knowledge needed to formulate informed plans. For instance, past successes or failures in similar situations, user preferences, or environmental observations stored in memory can guide the planning process~\cite{shinn2023reflexion}.
The agent's profile, defined by the profiling module, can also influence the style and focus of the planning process. For example, an agent profiled as risk-averse might prioritise plans with higher certainty of success, while a creative problem-solver profile might encourage the exploration of novel or less conventional plans~\cite{zhang2024agentic}.

\subsubsection{Action}

The primary goal of the Action Module is to translate the agent's formulated plans into specific outcomes within the environment. It acts as the execution engine, directly engaging with the surrounding world~\cite{wang2024survey}.
The intended outcomes of agent actions are diverse and intrinsically linked to the agent's designated task. Actions may be directed towards achieving concrete objectives, such as item crafting in a game~\cite{wang2023voyager} or software development task completion~\cite{wang2024openhands}, where each action contributes directly to the final goal. In the context of a web browsing agent~\cite{lai2024autowebglm}, actions can include navigating or interacting to specific pages. Actions may also facilitate information sharing and collaboration with other agents~\cite{gan2025master} or human users ~\cite{franciscatto2025cbr}, including conversational exchanges and feedback provision. Furthermore, agents perform exploratory actions to gain new knowledge and expand perception~\cite{wen2024autodroidv1}, optimizing learning and performance through exploration-exploitation. The Action Module receives from the Planning Module a sequence of steps or a specific action to be taken and translates them into executable actions~\cite{wang2024survey}. The Action Module's operational mechanism varies with the agent's architecture and action space, which collectively define the agent's interaction capabilities. Actions can be broadly categorised as follows:
\begin{itemize}
    \item \textbf{External Tool and API Interaction:} LLM-driven agents frequently generate natural language commands or structured calls to external tools and APIs~\cite{schick2023toolformer}. This allows them to interact with the external world, using resources like search engines for web navigation~\cite{lai2024autowebglm} (including clicking, form filling, and scrolling), databases via SQL queries~\cite{shen2024demonstration}, code execution environments~\cite{wang2024openhands}, PDDL solver execution~\cite{chang2024agentboard}, or specific application such as FlightSearch API with correctly formatted parameters~\cite{xie2024travelplanner}. To simplify the integration of diverse APIs, the Model Context Protocol (MCP)~\cite{unknown-author-2024} standardises API access, eliminating manual tracking and description. MCP's Host-Client-Server architecture facilitates dynamic tool discovery and execution, enabling any MCP-compatible LLM application to use connected server tools.
    \item \textbf{Internal Response Generation:} This category includes actions that generate internal responses, such as providing explanations or refining plans~\cite{yao2023react, shinn2023reflexion}. This is crucial as each action's consequence, affecting the user's information state or the environment's state, provides vital feedback. The Planning Module uses this feedback to iteratively adjust the agent's policy and refine future actions.
    \item \textbf{Embodied Navigation and Manipulation:} For embodied agents or robotic control systems, the Action Module grounds high-level skills or planned actions into low-level motor commands executable in physical or virtual environments~\cite{xi2025rise}. This serves as an intermediary, bridging the gap between cognitive planning and physical execution.
\end{itemize}
Therefore the design of a comprehensive and precise action space is crucial for building robust agents, as it standardises the translation of high-level plans into executable operations. To enhance execution accuracy, action names should semantically correspond to their behavior, and natural language descriptions should be provided to clarify action usage~\cite{xu2024crab}.

\subsubsection{Agent Topologies}
The design strategies employed for LLM Agents vary significantly based on the complexity and domain of the target tasks. They often involve augmenting the LLM core with specialized architectures for memory management, external knowledge access, and collaborative reasoning:
\begin{itemize}
    \item \textbf{Single Agent + action tools}: This strategy equips a single LLM agent with the ability to execute code or commands within a dedicated and isolated environment, typically referred to as a sandbox or a Graphical User Interface (GUI), which provide a set of tools for the interaction with operating systems or code generation~\cite{zhuge2024agent, shinn2023reflexion, lai2024autowebglm}. Usually a sandbox is provided for a secure execution space for arbitrary code generated by the agent, mitigating potential safety risks associated with deploying unpredictable LLM output directly into production systems. 
    
    \item \textbf{Agent + Information Retrieval (RAG, Tools, APIs)}: This approach focuses on overcoming the limitations of an LLM's static internal knowledge (hallucination, outdated facts) by augmenting its input context with external, non-parametric knowledge retrieved from diverse sources~\cite{li2025search, wu2024avatar, zhang2024agentic}.
    This approach focuses on overcoming the limitations of an LLM's static internal knowledge (hallucination, outdated facts) by augmenting its input context with external, non-parametric knowledge retrieved from diverse sources. This technique is used for factual and updated response such as in medical and enterprise domains.
    
    \item \textbf{Multi-Agent Systems (MAS):} This paradigm overcomes the cognitive bottlenecks of monolithic LLMs by distributing complex problem-solving across a network of specialized agents. Rather than relying on a single inference chain, MAS architectures enforce coordination and communication to achieve shared objectives. Interaction patterns typically fall into two categories: \textit{collaborative execution}, where agents assume distinct functional roles (e.g., Planner, Executor, Reviewer) to parallelize sub-tasks~\cite{erdogan2025plan, gebreab2024llm}; and \textit{iterative refinement}, where agents engage in debate or feedback loops to critique and improve outputs before final delivery~\cite{chan2023chateval, chen2025improving}. This topology is increasingly essential in high-stakes domains, such as clinical trial analysis~\cite{yue2024clinicalagent} and autonomous software maintenance~\cite{tao2024magis}.
    
\end{itemize}

\begin{table}[htbp]
\caption{Key LLM agent frameworks and benchmarks. The benchmarks are grouped by their primary domain focus to highlight differences in environments and key characteristics.}
\label{tab:agent-benchmarks}
\centering
\small
\begin{tabularx}{\columnwidth}{@{} l >{\raggedright\arraybackslash}X >{\raggedright\arraybackslash}X @{}}
\toprule
\textbf{Benchmark/Framework} & \textbf{Evaluation Focus} & \textbf{Environment} \\
\midrule
AgentBench~\cite{liu2023agentbench} & Multi-domain capabilities & OS, databases, knowledge graphs, web browsers.\\
\addlinespace

TheAgentCompany~\cite{xu2024theagentcompany} & Simulated professional software development work & Self-contained simulated corporate environment (OS, web tools, worker communications).\\
DevAI~\cite{zhuge2024agent} & AI application development tasks & Coding environment.\\
SWE-bench~\cite{yang2024swe} & Real-world software engineering maintenance & 12 Python GitHub repositories.\\
\addlinespace

WebShop~\cite{yao2022webshop} & E-commerce product search and retrieval & Simulated online shopping website.\\
WebArena~\cite{zhou2023webarena} & End-to-end general web task completion & Comprehensive, fully functional website sandboxes.\\
\addlinespace

ALFWorld~\cite{shridhar2020alfworld} & Interactive situated decision-making & Text-based game environment.\\
Minecraft~\cite{wang2023voyager} & Open-ended exploration and complex task execution & 3D voxel-based game world.\\
RoCoBench~\cite{mandi2024roco} & Multi-agent collaboration and coordination & Cooperative robotics scenarios.\\
\addlinespace

TravelPlanner~\cite{xie2024travelplanner} & Real-world constraint satisfaction planning & Information travel scenarios.\\
ScienceWorld~\cite{wang2022scienceworld} & Scientific reasoning and procedural planning & Simulated science laboratory environment.\\
\addlinespace

AgentDojo~\cite{debenedetti2024agentdojo} & Safety and adversarial resilience & Adversarial prompt injection setups.\\
PrivacyLens~\cite{shao2024privacylens} & Privacy norm adherence and data protection & Scenarios involving sensitive data handling.\\
\bottomrule
\end{tabularx}
\end{table}

\subsection{Agent Evaluation}

Evaluating the capabilities of AI agents is crucial for their development and integration into various applications and specific domains. As the field of LLM-based autonomous agents grows, the need for robust evaluation methods becomes important. These tools allow for rigorous testing of architecture effectiveness within the intended operational domain. 
Generally, agent evaluation can be categorised into two main approaches: subjective and objective.

\subsubsection{Subjective Evaluation}
Subjective evaluation is particularly effective when assessing qualitative aspects of agent performance, such as overall helpfulness or user-friendliness, where quantitative metrics are difficult to define or evaluation datasets are limited. LLM-based agents are often designed to serve humans, making subjective evaluation a critical component as it reflects human criteria.
This involves human evaluators directly interacting with the system or observing its performance, and then providing judgments based on predefined criteria or overall impressions. This approach often uses structured questionnaires with Likert scales to assess qualitative aspects. For instance, Franciscatto et al.~\cite{franciscatto2025cbr}  employed a 5-point Likert scale to evaluate visibility, support, usefulness, transparency, justification, and data integration capabilities. Similarly, Shaer et al.~\cite{shaer2024ai} used Likert scales to assess generated ideas for relevance, innovation, and insightfulness by both expert and novice evaluators. Furthermore, in~\cite{schmidgall2025agent} the authors conducted human surveys with PhD researchers, who rated paper quality on experimental quality, report quality, and usefulness, and also simulated peer review using NeurIPS-style scores.
Subjective evaluation, while offering valuable qualitative insights, shows a series of considerable challenges. Primarily, it is resource-intensive and time-consuming, necessitating the recruitment of a sufficiently large and diverse pool of human evaluators, which can incur substantial financial costs and require extensive logistical planning for study preparation and execution. Moreover, the inherent subjectivity of human judgment introduces potential bias, influenced by individual perspectives, experiences, and cultural backgrounds. These biases can disrupt the objectivity and generalizability of evaluation results. Consequently, the interpretation and integration of subjective evaluation data require careful consideration of these inherent limitations, emphasizing the need for rigorous methodological design and transparent reporting to mitigate potential biases and enhance the reliability and validity of findings. 
\subsubsection{Objective Evaluation}
Objective evaluation uses quantitative metrics to enable computational, and comparative analysis, thereby providing concrete and measurable assessments of agent performance. The selection of appropriate metrics is crucial for evaluating specific aspects of agent performance. For \textbf{Task Completion and Efficiency}, evaluation focuses on the agent's ability to achieve defined objectives with efficient resource use. This category includes metrics such as Success Rate~\cite{chang2024agentboard}, quantifying overall goal attainment; Progress Rate~\cite{chang2024agentboard}, tracking incremental advancements in multi-turn interactions; Completion Rate~\cite{xu2024crab}, measuring the proportion of completed sub-tasks; Execution Efficiency~\cite{xu2024crab}, assessing action efficiency relative to sub-task completion; Cost Efficiency~\cite{xu2024crab}, evaluating resource consumption through token usage; Number of Steps~\cite{xu2024theagentcompany}, quantifying operational effort via LLM calls; and Cost per Instance~\cite{xu2024theagentcompany}, measuring monetary expenditure for API queries. In numerous applications, \textbf{Accuracy and Correctness} have critical importance. Metrics within this category assess the fidelity and consistency of generated information, plans, and actions. This includes Accuracy~\cite{wang2024survey}, evaluating overall output correctness; Answer F1~\cite{liu2023agentbench}, measuring query response accuracy in knowledge graph environments; Test Coverage and Bug Detection Rate~\cite{wang2024survey}, assessing the agent's ability to generate effective test cases and identify software defects. For agents designed for conversational or interactive tasks, \textbf{Dialogue and Interaction Quality} is assessed using metrics such as Recency, Relevance, and Importance~\cite{hou2024my}. These metrics evaluate the coherence, relevance, and meaningfulness of agent dialogues, capturing the nuances of human-agent interaction. Finally, as AI agents are increasingly deployed in critical systems, \textbf{Adversarial Robustness} becomes a key evaluative consideration. Metrics such as Benign Utility, Utility Under Attack, and targeted Attack Success Rate~\cite{debenedetti2024agentdojo} evaluate the agent's resilience to adversarial attacks, ensuring security and reliability in real-world deployments.

\paragraph{Frameworks and Benchmarks} To comprehensively evaluate the capabilities of agents across diverse application domains, a range of specialised benchmarks and frameworks have been developed. These include AgentBench ~\cite{liu2023agentbench} , which features tasks simulating real-world scenarios such as web browsing and gaming, alongside code generation and execution tasks involving operating systems, databases, and knowledge graphs. TheAgentCompany~\cite{xu2024theagentcompany} provides a self-contained environment that models a small software company, including tasks like web browsing, code writing, and inter-agent communication. DevAI~\cite{zhuge2024agent} presents 55 real-world AI application development tasks curated by expert annotators, while SWE-bench~\cite{yang2024swe} offers 2294 software engineering problems derived from actual GitHub issues and pull requests across 12 prominent Python repositories. Text-based game environments, such as ALFWorld~\cite{shridhar2020alfworld}, and Minecraft~\cite{wang2023voyager}, are used to evaluate language agent performance in interactive, simulated settings. WebShop~\cite{yao2022webshop} focuses on assessing product search and retrieval capabilities, and WebArena~\cite{zhou2023webarena} provides a comprehensive website environment for end-to-end agent evaluation. RoCoBench~\cite{mandi2024roco} evaluates multi-agent collaboration across diverse scenarios, emphasizing communication and coordination in cooperative robotics. TravelPlanner ~\cite{xie2024travelplanner} benchmarks real-world planning capabilities in language agents, and ScienceWorld ~\cite{wang2022scienceworld} evaluates reasoning and planning abilities by posing questions designed to challenge a fifth-grade student. Finally, for evaluating safety and privacy, AgentDojo ~\cite{debenedetti2024agentdojo} measures AI agent resilience to prompt injection attacks, and PrivacyLens ~\cite{shao2024privacylens} quantifies potential data leakage to assess privacy norm adherence in language model agents. All these framework and benchmarks are summarised in Table \ref{tab:agent-benchmarks}.

\subsubsection{LLM-Evaluators}
A growing number of researchers are also exploring the use of Large Language Models (LLMs) themselves as intermediaries for agent assessment. For instance, ALI-Agent~\cite{wang2024ali} leverages autonomous LLM-driven agents to automatically generate realistic test scenarios and iteratively refine them, enabling adaptive evaluations of alignment that effectively identify subtle, long-tail risks without relying on continuous human feedback. This framework employs a memory module for scenario generation, a tool-using module that integrates Web search and fine-tuned evaluators to reduce human labor, and an action module for test refinement.
Building upon the concept of LLMs as judges, the Agent-as-a-Judge framework~\cite{zhuge2024agent} extends this by employing agents to evaluate other agents. Recognizing the step-by-step operation of the agents, this framework aims to provide rich, intermediate feedback throughout the task-solving process, rather than relying solely on final outcomes. This approach has demonstrated success in code generation tasks, outperforming traditional LLM-as-a-Judge methods and achieving reliability comparable to human evaluations. ChatEval~\cite{chan2023chateval} employs a multi-agent debate approach to enhance LLM-based evaluation quality. By deploying a team of LLM agents with diverse role prompts, ChatEval facilitates autonomous discussion and evaluation of generated responses. This synergistic approach, leveraging the unique capabilities of multiple LLMs, exhibits superior accuracy and correlation with human assessments compared to single-agent evaluations. Similar multi-agent systems, such as IntellAgent~\cite{levi2025intellagent}, use an LLM to perform roles like event generation, user agent, and dialog critique, implicitly integrating LLMs into the evaluation of agent behavior.
SCALEEVAL~\cite{chern2024can} proposes an agent-debate-assisted meta-evaluation framework to address this, employing communicative LLM agents for iterative discussions that assist human annotators in identifying the most capable LLM evaluators, particularly in novel, user-defined scenarios. In cases of agent disagreement, minimal human oversight ensures a balance between efficiency and reliability. This approach enables scalable assessment of LLM evaluator trustworthiness across diverse tasks and criteria.
While LLM-based evaluation offers advantages such as automation, scalability, and the ability to probe complex behaviors, it also presents challenges. These include reliance on inherent LLM capabilities and potential biases, which may cause incorrect evaluations, diverging from expert human judgment in domain-specific tasks~\cite{szymanski2025limitations}. Furthermore, LLM evaluations are susceptible to prompt engineering and format variations, potentially leading to inconsistent assessments~\cite{chern2024can}. The inherent lack of understanding in specialised domains can result in potential errors that human experts would make~\cite{szymanski2025limitations}.

\subsection{Field of Use}
The domain-specific nature of agent evaluation highlights their adaptability, as reflected in the diverse metrics and frameworks used across various sectors. LLM-based agents demonstrate broad applicability in domains such as:
\begin{itemize}
    
    \item \textbf{Software Engineering and Code Generation:} Agents in this domain have moved beyond simple code completion to handling full development lifecycles. Current research focuses on automating distinct engineering tasks. First, agents perform secure \textit{environment interaction} and \textit{workflow simulation}; for instance, platforms like OpenHands~\cite{wang2024openhands} execute scripts and commands in sandboxed environments to mimic human developer workflows. Second, agents tackle \textit{iterative debugging} and \textit{self-correction}. For instance Reflexion~\cite{shinn2023reflexion} uses verbal feedback and episodic memory to reason through and fix code errors or MAGIS~\cite{tao2024magis}, which employs a multi-agent framework to resolve GitHub issues, implement feature updates, and perform code optimizations.
    
    \item \textbf{Embodied Agents:} In physical and simulated environments, agents are tasked with grounding high-level language into low-level control policies. Recent approaches leverage foundation models to bridge this gap across distinct domains. In \textit{robotic control}, Magma~\cite{yang2025magma} introduces spatial-temporal reasoning via Set-of-Mark and Trace-of-Mark to bridge verbal and spatial intelligence, while LAC~\cite{xu2024lac} employs a multi-agent controller framework that merges LLM reasoning with feedback control loops for real-time responsiveness. For \textit{open-ended exploration}, agents are used in Minecraft sandboxes: Voyager~\cite{wang2023voyager} demonstrates lifelong learning through an automatic curriculum and an executable code skill library, whereas Steve~\cite{zhao2024see} enhances planning by integrating visual perception components with language instruction. Finally agents can be used in \textit{web task execution}. For instance WebPilot~\cite{zhang2024webpilotversatileautonomousmultiagent} addresses dynamic environments through a dual-strategy Monte Carlo Tree Search (MCTS), while AutoWebGLM~\cite{lai2024autowebglm} uses simplified HTML representations and reinforcement learning to navigate complex real-world webpages effectively.

    \item \textbf{Research and Development:} Agents are increasingly tasked with automating distinct phases of the scientific method. For \textit{lifecycle management}, frameworks like Agent Laboratory~\cite{schmidgall2025agent} act as interactive co-pilots, coordinating specialized sub-agents to handle literature reviews, experimentation, and report generation. In the field of \textit{autonomous discovery}, the AI Scientist Agent~\cite{lu2024ai} pushes boundaries by generating novel ideas, executing code-based experiments, and authoring full manuscripts without human intervention. Also ADAS~\cite{hu2024automated} iteratively programs and optimizes new agentic architectures in code space, discovering novel building blocks with cross-domain transferability.

    \item \textbf{Information Management and Retrieval:} Agents go beyond static information extraction by autonomously optimizing when and how to retrieve and process data. They are increasingly used for \textit{collaborative pipeline optimization}, where individual RAG components (e.g., query rewriters, selectors) operate as cooperative agents trained via Multi-Agent Reinforcement Learning to maximize global answer quality~\cite{chen2025improving}. Additionally, agents facilitate \textit{dynamic knowledge reasoning}, exemplified by frameworks that autonomously trigger searches during chain-of-thought processes and employ dedicated modules to refine retrieved documents into logic-compatible insights~\cite{li2025search}.

    \item \textbf{Healthcare and Medicine:} In critical safety domains, agents automate complex clinical and administrative tasks. For \textit{clinical trial optimization}, agents like ClinicalAgent~\cite{yue2024clinicalagent} utilize external biomedical knowledge graphs to predict trial outcomes by autonomously assessing drug efficacy, safety risks, and enrollment feasibility. Agents are also used for \textit{hospital administration}, for istance multi-agent frameworks~\cite{gebreab2024llm} streamline operations by autonomously navigating Electronic Medical Record (EMR) platforms to execute tasks such as patient registration, billing, and appointment scheduling.

\end{itemize}
The versatility of LLM-based agents demonstrated across various domains motivates their exploration within everyday environments like smartphones and IoT devices. Their deployment in pervasive computing settings offers solutions for complex decision automation, enhanced user interaction, and improved accessibility. This potential is supported by recent advances, including compact and efficient LLMs~\cite{qwq32b, yang2025magma}, increased memory in embedded systems, power-efficient processors, specialised hardware such as NPUs, and high-speed communication technologies like 5G.
The following sections explore the pervasive computing landscape, the evolving role of AI within it, and the integration of LLM-based agents in pervasive environments.

\section{Pervasive Computing: an Overview}
\label{sec:pervasive}
Pervasive computing, also referred to as ubiquitous computing, was conceived in early seminal work as technology that would 'weave themselves into the fabric of everyday life until they are indistinguishable from it'~\cite{weiser1991computer}. This concept focuses on the creation of environments densely populated with computing and communication capabilities, yet seamlessly integrated with human users, until the technology effectively becomes "transparent"~\cite{satyanarayanan2001pervasive, saha2003pervasive}. So while distributed systems and mobile computing enabled "anytime, anywhere" access to information, though not always with guaranteed connectivity, pervasive computing fundamentally shifts towards an "all the time, everywhere" presence, ensuring seamless access to computing whenever and wherever it is needed~\cite{saha2003pervasive}. Pervasive computing manifests in various devices, formats, and locations, ranging from resource-constrained sensors to high-performance servers, including cloud datacenters, mobile edge computing servers, mobile devices, TVs, wearables, sensors and embedded systems. These interconnected devices leverage various wireless communication technologies to enhance their capabilities while minimizing resource consumption, including battery power, memory, and CPU time leading to the proliferation of Internet of Things (IoT) devices. These IoT devices, endowed with sensing, computing, networking, and communication functionalities, are capable of collecting, analysing, and transmitting a diverse spectrum of data, including images, videos, audio, texts, wireless signals, and physiological signals from individuals and the physical environment. Furthermore sensors, integral to many IoT devices, play a crucial role by monitoring phenomena both within and beyond human perception, converting physical occurrences into numerical data. The size reduction, increased efficiency, enhanced sensitivity, and improved connectivity of these sensors have enabled their embedding in diverse environments and objects, providing real-time data and insights previously inaccessible. Therefore this increase in connectivity, data generation and accessibility of sensory data, contributes to the generation of zettabytes of real-time data streams~\cite{baccour2022pervasive} that can be used to create much more customizable and accurate systems but also can be challenging to manage. These pervasive systems find application across diverse sectors, ranging from Smart Homes and Smart Cities to Smart Factory and Healthcare. Cisco's projections\footnote{\href{https://www.cisco.com/c/en/us/products/collateral/wireless/iot-mobility-services/5g-iot-services-wp.pdf}{Cisco 5G IoT White Paper}} indicate over 2.6 billion cellular-connected IoT devices by 2026, showing the maturation of their capabilities and decreasing costs. Furthermore, the economic impact of IoT is projected to be substantial, with estimates suggesting a global value creation ranging from \$5.5 trillion to \$12.6 trillion by 2030~\cite{chui2021internet}. All these factors underscore the substantial and growing impact of pervasive computing environments in improving the quality of life.

\subsection{Computational and Network Advancements}
The increasing diffusion of pervasive computing is significantly driven by the advancements in both computational power and network technologies.
Embedding computational power within everyday objects is fundamental to pervasive computing. This necessitates a range of microprocessor solutions tailored to specific device requirements. These range from low-power Microcontroller Units (MCUs) for basic sensing and actuation in resource-constrained devices, to more powerful Central Processing Units (CPUs) for complex processing tasks, and highly integrated Systems on a Chip (SoCs) for feature-rich applications. SoCs often incorporate specialised processing units such as Graphics Processing Units (GPUs) and Digital Signal Processors (DSPs) to enhance multimedia and signal processing capabilities within compact form factors. Notably, the emergence of architectures like PlasticARM CPUs~\cite{biggs2021natively} presents a promising avenue for pervasive devices, including smartphones and edge computing nodes, offering an optimised balance of computational performance and energy efficiency crucial for their operation. This continuous evolution in microprocessor technology allows for increasingly sophisticated computational intelligence to be integrated directly into the fabric of everyday life. Also the increasing integration of Artificial Intelligence into edge devices within pervasive computing environments demands dedicated processing capabilities for computationally intensive AI tasks. Recognizing the limitations of general-purpose CPUs and GPUs for these specialised workloads, the development of dedicated AI accelerators has been a critical enabler. These specialised silicon solutions, often implemented as discrete co-processors or integrated within SoCs as Neural Processing Units (NPUs)~\cite{tan2021efficient}, offer significantly enhanced performance and energy efficiency for neural network computations. The strategic inclusion of NPUs in SoCs designed for IoT, edge computing, and mobile platforms facilitates the rapid and efficient execution of complex AI algorithms directly on the device, minimizing latency and power consumption associated with cloud-based processing. 
The seamless operation of interconnected devices, a defining characteristic of pervasive computing, relies critically on robust and efficient networking infrastructure. Given the heterogeneity of devices and their diverse communication requirements, a suite of networking technologies is essential. While high-bandwidth applications benefit from advancements in Wi-Fi standards like Wi-Fi 7 with its focus on minimal latency, low-power devices leverage technologies such as ZigBee for personal area networks. Wide-area connectivity for distributed IoT deployments is facilitated by Low Power Wide Area Networks (LPWANs) like LoRaWAN~\cite{bimpas2024leveraging}. Furthermore, high-speed cellular technologies like 5G and 6G provide ubiquitous connectivity for mobile and edge devices. This diverse and evolving network infrastructure provides the essential connectivity for the complex communication and coordination within pervasive computing environments.


Pervasive computing leverages various architectural infrastructures to deliver its capabilities, primarily involving the edge-to-cloud continuum paradigm (Figure~\ref{fig:pervasive_layer}). The cloud layer offers high computational power and storage for complex tasks and centralised management. The edge layer, situated closer to the user and the data, facilitates distributed, latency-aware applications by providing local computing, network connectivity and decision-making at the data source for real-time interactions. These architectures work in a complementary way to enable the collection, processing, and use of data generated by a multitude of interconnected devices.

\subsubsection{Cloud Layer}

As defined by the National Institute of Standards and Technology (\textit{NIST}), cloud computing is '\textit{a model for enabling ubiquitous, convenient, on-demand network access to a shared pool of configurable computing resources (e.g., networks, servers, storage, applications, and services) that can be rapidly provisioned and released with minimal management effort or service provider interaction}'~\cite{mell2011nist}. This paradigm includes centralised or distributed computing technologies operating over the Internet, primarily functioning as a scalable storage and processing infrastructure. Parallel and distributed computing models can be independently or jointly integrated and deployed within data centres, either physically or as virtualised resources.
Within the context of pervasive computing, the cloud serves as a high-capacity processing hub for the extensive datasets generated by edge devices. Its inherent computational resources and scalability are paramount for managing high data flood~\cite{yalli2024internet}. Cloud service models are commonly categorised into private, community, public, and hybrid deployments~\cite{mell2011nist}. Private clouds are dedicated to a single organization, regardless of who manages it or its location. Community clouds serve a specific group with shared needs, potentially managed by members or a third party, on or off-site. Public clouds are openly accessible to anyone and are owned and operated by providers on their premises. Hybrid clouds combine two or more distinct cloud types, linked by technology allowing data and application movement between them.
Thanks to cloud computing, the connectivity of IoT devices to the cloud, and their integration with other related sensors, unlocks significant potentials: 
\begin{itemize}
    \item \textbf{Scalable Storage and Processing:} Cloud services provide the large-scale storage and high-performance distributed computing resources essential for managing the substantial data volumes generated by pervasive devices, effectively overcoming their inherent storage and processing limitations.
    \item \textbf{Stable Middleware Layer:} The cloud can establish a more robust middleware layer within the IoT architecture (positioned between IoT devices and applications) by centralizing service implementations.
    \item \textbf{Convenient and Cost-Effective Data Storage:} The convenience and economic benefits of cloud computing have driven widespread adoption by individuals and enterprises for storing data originating from pervasive devices.
\end{itemize}
Fundamentally, cloud computing signifies a transition from traditional, localised computing paradigms to a model characterised by flexible resource sharing and reduced operational costs. However, relying solely on cloud computing also presents challenges:
\begin{itemize}
    \item \textbf{Latency Issues:} Network transmission delays or extensive response queuing can introduce significant latency, potentially hindering the performance of real-time applications.
    \item \textbf{Privacy Concerns:} Centralizing data in the cloud and transmitting sensitive personal information over the internet raises substantial privacy concerns. Users must be cognizant of the risks associated with cloud storage, including the security of their private data and the potential for data breaches.
    \item \textbf{High Bandwidth Costs:} Transmitting the large volumes of diverse data (text, video, images, audio, and IoT sensor readings) generated by pervasive devices to cloud data centers can incur substantial bandwidth costs and strain network infrastructure.
\end{itemize}

\subsubsection{Edge Layer}
This is the pivotal layer of the pervasive architecture, performing computations closer to the data source to enhance responsiveness and efficiency. As defined by the National Institute of Standards and Technology (NIST), edge computing '\textit{is the network layer including the end-devices and their users, to provide, for example, local computing capability on a sensor, metering or some other devices that are network-accessible}'~\cite{iorga2017nist}. 
The edge layer has a wide variety of heterogeneous devices, spanning from the mere sensors and actuators, to \textbf{resource-constrained devices} (e.g., microcontrollers) capable of basic logic, up to \textbf{powerful user-end units} (e.g., smartphones, PCs) that can provide high computational capabilities~\cite{sun2016edgeiot, baccour2022pervasive}. By decentralizing intelligence and processing power, this paradigm leverages this entire spectrum situated in close proximity to the data-generating entity~\cite{yalli2024internet}.

To bridge the gap between these distributed edge devices and the centralized cloud, the architecture incorporates an intermediate infrastructure layer, historically referred to as \textbf{Fog Computing}. Functioning as a high-performance tier within the edge continuum, these nodes reside physically between smart end-devices and centralized services to facilitate connectivity and heavy processing. As defined by NIST, this infrastructure '\textit{facilitates the deployment of distributed, latency-aware applications and services, and consists of fog nodes (physical or virtual), residing between smart end-devices and centralised services}'~\cite{iorga2017nist}. These nodes include physical components such as Raspberry Pi, NVIDIA Jetson platforms, gateways, routers, or nano-servers, which are capable of handling larger volumes of data than simple sensors while avoiding the latency of the cloud~\cite{sabireen2021review, yalli2024internet}. This intermediate tier usefull for offloading computations from resource-constrained edges, addressing bandwidth and energy consumption problems by processing data locally to produce high-quality results before transmission.

A specific implementation of the Edge Layer is Mobile Edge Computing (MEC), which deploys services and computational capabilities at the edge of cellular networks, close to subscribers. This strategic placement enables a service environment characterised by ultra-low latency, high bandwidth, and direct access to real-time network information~\cite{li2016mobile}. For instance, in~\cite{sun2016edgeiot}, researchers handle data streams at the mobile edge to overcome the scalability limits of traditional IoT architectures, reducing traffic load in the core network.

However, implementing and managing this heterogeneous edge environment presents several challenges:
\begin{itemize}
    \item \textbf{Device and Network Management:} Due to its decentralised and heterogeneous nature, managing randomly distributed network resources and high-risk device breakdowns complicates connectivity and application deployment.
    
    \item \textbf{Computational Challenges:} The hierarchical structure of edge systems makes optimal task allocation across IoT devices, intermediate nodes, and the cloud challenging. Ensuring computational correctness in such distributed environments is complex, further compounded by the need to choose suitable protocols for heterogeneous sensors and devices.
    
    \item \textbf{Security Challenges:} Heterogeneous devices in less secure physical environments can be susceptible to various attacks, including man-in-the-middle attacks. Furthermore, while processing sensitive data locally enhances privacy by minimizing cloud transfer, it requires robust local security measures to maintain user trust.
\end{itemize}

\section{Pervasive AI Computing}

The integration of Artificial Intelligence into the pervasive computing domain is driven by its strong capability to resolve complex tasks that traditional computational methods cannot handle efficiently. From classifying intricate patterns in images and text to generating sophisticated sequences of actions, AI transforms various heterogeneous environmental inputs into actionable intelligence. This integration is significantly enabled by the rapid advancements in computational hardware, which allow the deployment of AI models even on resource constrained devices. This convergence of advanced problem-solving capabilities and hardware evolution has established the field of Pervasive AI~\cite{baccour2022pervasive} defined as the '\textit{intelligent and efficient distribution of AI tasks and models over/among any types of devices with heterogeneous capabilities in order to execute sophisticated global missions}'.
This paradigm marks a departure from traditional, cloud centralised AI approaches, leveraging the distributed computational resources inherent in pervasive environments, including IoT devices and edge servers. Within this domain, AI-enabled sensors play a crucial role, categorised as: AIoT Sensors, facilitating cloud-based AI decision-making based on physical-world data; Edge AI Sensors, enabling localised AI inference at the device level; and TinyML Sensors, designed for efficient execution of specific tasks with minimal data~\cite{sloane2025materiality}.
In this field, Deep Learning plays a key role in the diffusion and application of AI models within pervasive computing. 

\subsubsection{Deep Learning}
Deep Learning, an important subfield of Machine Learning that takes inspiration from biological nervous systems. It uses deep neural networks (DNNs), which are characterised by a high number of interconnected layers of neurons that learn features to accomplish a task. The multilayer perceptron, for instance, consists of fully connected neurons employing nonlinear activation functions. In contrast, convolutional neural networks (CNNs), prevalent in vision tasks, use convolutional layers. Each convolutional layer incorporates a set of learnable parameters, called filters, which possess the same number of channels as the input feature maps but with smaller spatial dimensions. Each filter channel convolves across the length and width of its corresponding input feature map, computing the inner product. The summation of these channel-wise inner products yields a single output feature map, and the total number of output feature maps corresponds to the number of applied filters. Another prominent architecture is the Transformer, which processes text or images encoded as vector embeddings (tokens).  
These models support a range of applications in autonomous systems, robotics, smart homes, and virtual reality. However, deploying DL models on resource-constrained edge devices requires balancing accuracy with efficiency.
To address the challenges of deploying deep learning on resource-constrained pervasive devices, various strategies are being explored such as: 
\begin{itemize}
    \item \textbf{Distributed inference:} Splits DL models across devices to reduce local load and cloud latency, e.g. EdgeShard~\cite{zhang2024edgeshard}.
    \item \textbf{Federated learning (FL):} Trains models across devices while keeping data local, supporting privacy and scalability~\cite{qu2025mobile}.
    \item \textbf{Optimization techniques:} Include caching, compression, and dynamic inference to improve DNN performance on constrained devices~\cite{siam2025artificial}.
\end{itemize}

While traditional DL models have empowered pervasive devices to effectively perceive complex sensory data (e.g., computer vision, text processing), they operate as rigid inference engines. These models typically depend on strictly structured inputs and are confined to a pre-determined set of outputs, effectively mapping specific signals to fixed labels or commands. The novelty of Agentic AI lies in the integration of Large Language Models (LLMs), which fundamentally disrupts this paradigm. Unlike traditional models, LLMs can interpret complex, unstructured natural language queries without requiring a rigid input format. They possess the ability to generalize user intent, infer logical plans, and generate dynamic action sequences to resolve novel tasks. This shift advances pervasive systems from merely classifying environmental states to actively reasoning about them, thereby enabling autonomous problem-solving in dynamic environments.

The next section explores how these LLM-based agents are specifically applied in pervasive computing and addresses the significant challenges of deploying such powerful reasoning engines on limited-resource platforms.

\section{Agents in Pervasive Computing}
\label{sec:pervasiveagent}
The integration of intelligent agents within pervasive computing environments represents a significant advancement in human-computer interaction, leading to more intuitive and proactive experiences. Pervasive computing, characterised by the integration of embedding of computational capabilities into everyday objects and environments, provides an ideal environment for the deployment of agents capable of perceiving, reasoning, and acting autonomously.
As discussed in Section~\ref{sec:agents}, the advent of Large Language Models (LLMs) has substantially augmented the potential of these agents by equipping them with advanced natural language understanding and generation capabilities, enabling their application in a wide range of use cases. This enhancement facilitates and increase the performance of this pervasive devices with more intuitive and natural interactions within smart environments, such as automatic voice-controlled smart homes~\cite{rivkin2024aiot}, automatic smartphone interactions~\cite{wen2024autodroidv1}, or automatic traffic management in smart cities~\cite{chen2024llm}.
However, the deployment of LLM-powered agents in pervasive computing scenarios presents various challenges. As discussed in Section~\ref{sec:pervasive}, one of the primary concern arises from the resource constraints inherent in many pervasive devices, including limitations in processing power and energy availability. This contrasts with the considerable computational, energy, and memory storage demands of LLMs. Moreover, the direct interaction of these agents with user data on pervasive devices underscores the critical importance of user data privacy. Consequently, ensuring both the sustainable operation of these agents and the robust safeguarding of user data are essential for their responsible development and deployment. Addressing these challenges necessitates research into several key areas. This includes determining optimal deployment and alignment strategies for LLMs in both local and distributed pervasive environments. Also, research must focus on redesigning the core components of agents specifically memory, planning, reasoning, and acting modules to operate efficiently within the resource limitations inherent in pervasive computing. Furthermore, rigorous evaluation of agent performance within these specific pervasive environments is crucial to identify their strengths and weaknesses.

To address these challenges, this section provides a comprehensive analysis of the adaptation of agents for pervasive environments. First, we examine the deployment strategies of the LLMs, evaluating the trade-offs between local execution and distributed (edge/cloud) architectures. Next, we discuss the alignment strategies, focusing on techniques like quantization and fine-tuning that tailor models to resource-constrained domains. We then explore the agent design strategies, detailing how core the core modules (memory, planning, and action) are adapted for pervasive constraints. Finally, we review real-world evaluation metrics and applications across various pervasive environments such as smart homes, mobile devices, and healthcare.

\subsection{LLMs Deployment Strategies}
\label{sec:deploym}
Addressing the integration of LLMs into resource-constrained pervasive computing environments involves in two primary deployment strategies: local and distributed.
The local deployment strategy aims to execute the LLM directly on the device where the user data is stored, minimizing latency and enhancing data privacy. However, this approach is constrained by the limited computational and resource capabilities of the local device.
The distributed deployment strategy aims to augment computational resources through cloud computing servers, which offer enhanced processing power but introduce greater latency and raise privacy issues. Alternatively, this strategy involves partitioning the Large Language Model across multiple fog or edge devices, a method that strategically locates processing closer to the user's device to decrease latency, but increases system complexity.

\subsubsection{Local LLM Deployment}
Local deployment of LLMs aims to execute these powerful models directly on resource-constrained edge devices, offering significant advantages in terms of user privacy, cost-effectiveness by eliminating reliance on cloud infrastructure, and reduced latency due to on-device processing. However, the inherent limitations in computational power, energy availability, and storage capacity of these pervasive devices present substantial challenges to implement this vision. A key strategy to overcome these limitations involves optimizing LLM resource consumption through techniques such as small language models (SLMs) deployment, model quantization (to reduce the bit-precision of model weights), and pruning techniques (to reduce model parameters).
For instance, researchers in~\cite{wen2024autodroidv1} successfully implemented an agentic architecture using a Vicuna-7B, on a OnePlus ACE 2 Pro smartphone equipped with a Snapdragon 8 Gen2 CPU and Adreno™ 740 GPU. Building upon this, Autodroid v2~\cite{wen2024autodroid} demonstrated impressive performance in success rate and inference latency on the same resource-constrained device by deploying an 8-bit quantised Llama3.1-8B model. Further highlighting the potential of smaller models, Microsoft's Magma~\cite{yang2025magma}, an 8B (where 'B' denotes Billion) parameter model, exhibited strong verbal and spatial-temporal reasoning in UI navigation and robotic manipulation tasks. Additionally, the MaViLa framework~\cite{fan2025mavila} showcased the effectiveness of smaller models by fine-tuning a Vicuna-13B model for smart manufacturing, achieving high performance in domain-specific tasks like additive manufacturing monitoring, anomaly detection, and autonomous process optimization. Despite the relative smallness of 7B or 13B parameter models compared to their larger 30B+ counterparts, their deployment on resource-constrained devices like typical smartphones (with 6-12 GB of RAM) remains a significant obstacle. As previously discussed, the feasibility of local LLM execution is heavily influenced by the model's parameter count and the memory footprint of the Key-Value (KV) cache. A Llama-2 7B model, for example, can require up to 28 GB for full precision inference, 7 GB with 8-bit quantization, and 3.5 GB with even 4-bit quantised versions, in addition to over 2 GB potentially needed for its 4k token context window's KV cache. While deployment might be viable on more powerful edge devices like NVIDIA Jetson Orin series~\footnote{\href{https://www.nvidia.com/it-it/autonomous-machines/embedded-systems/jetson-orin/}{Jetson Orin Technical Specification}} for developer and industrial applications, running such models efficiently alongside their KV cache on devices with less computational power, necessitates innovative memory management strategies. Furthermore, energy consumption is a critical concern, with research indicating an approximate cost of 0.1 J/token per billion parameters~\cite{liu2024mobilellm}. This suggests that a 7B parameter LLM could consume 0.7 J/token, potentially limiting continuous conversational usage. For instance a fully charged iPhone, with approximately 50 kJ of energy, can sustain this model in conversation for less than 2 hours at a rate of 10 tokens/s, with every 64 tokens draining 0.2\% of the battery~\cite{liu2024mobilellm}. This highlights the urgent need for techniques that significantly enhance the energy efficiency of these models.
To address these challenges, the literature presents various promising models and methodologies. Nexa AI's OmniVLM~\cite{chen2024omnivlm}, a compact model with under one billion parameters (968M), directly tackles the memory footprint and computational demand issues, making it more easy to deploy. OmniVLM also introduces a novel token compression mechanism for visual inputs, achieving a substantial reduction in visual token sequence length, thereby lowering computational overhead while preserving visual-semantic information.
Octopus v2~\cite{chen2024octopusv2} uses another approach, enabling a 2 billion parameter on-device language model to outperform GPT-4 in function calling accuracy and latency while drastically reducing context length. This significantly enhances the feasibility of deploying AI agents directly on edge devices without cloud reliance. Building on this success, Octopus v3~\cite{chen2024octopusv3} introduces a sub-billion parameter multimodal model capable of efficiently processing both visual and textual inputs using functional tokens and CLIP-based image encoding.
Beyond model optimization, innovative deployment and runtime techniques are being explored. EdgeLLM~\cite{yu2024edge} proposes a layerwise unified compression (LUC) method for dynamic pruning and quantization of LLM layers, coupled with adaptive layer tuning and voting. This approach achieves significant reductions in computation and memory demands, enabling fine-tuning and updating of LLMs even on smartphones with substantial speedups and reduced memory usage.
Another strategy, explored in~\cite{yin2024llm}, involves hosting a single, stateful LLM as a system service within the mobile operating system, accessible to applications via system APIs. This design minimizes memory duplication and supports persistent context management through chunk-wise KV cache compression and tolerance-aware memory management. Context switching is accelerated via a swapping-recompute pipeline that overlaps I/O and computation. The system employs fine-grained memory eviction strategies, such as LCTRU (Least Compression-Tolerable and Recently-Used queue), adapted to the compression sensitivity of different parts of the model.
Finally, EdgeMoE~\cite{yi2025edgemoe} faces the challenge of deploying extremely large and sparse Mixture-of-Experts (MoE) LLMs on mobile hardware. By treating device memory as a smart cache and selectively preloading only the most likely needed experts, EdgeMoE can execute models with over 10 billion parameters on small edge devices with minimal overhead. Its key innovations include expert-wise bitwidth adaptation and predictive expert caching, enabling real-time inference without exhausting device resources.

\subsubsection{Distributed LLM Deployment}
While various models and strategies exist for deploying LLMs directly on the edge device, interacting with users and data-generating sensors, this approach is often constrained by the necessity of using smaller, less performing models, and by the latency of the models on producing each token at inference time due to the hardware limitations. A key alternative to mitigate this issue is using a distributed approach. This strategy involves hosting the models either on cloud servers, which offer high computational and memory resources, or distributing the model's computation across multiple edge or fog nodes.

\paragraph{Cloud Distribution}
Cloud computing offers the capability to host large open-source or proprietary LLMs on powerful servers, often accessed via Web Applications or APIs provided by major companies such as OpenAI (offering models like GPT-4, GPT-4o, o1, and o3), Anthropic (providing models like Claude 2.1, Claude 3.5 Sonnet, Claude 3.5 Haiku, and Claude 3.7), or Mistral (providing models like Mistral Large and Mistral Small).
In this architecture, edge devices transmit user prompts to these remote servers hosting the LLMs. The server receives the prompt, processes it with the designated model, generates a response, and then sends the answer back to the originating edge device.
Cloud computing proves advantageous when high-performance LLMs are required. However, as the servers and models are typically managed by external entities, careful consideration must be given to user privacy, the potential for connection instability, and response latency.
Examples of cloud-based LLM applications include~\cite{rivkin2024aiot}, where models like GPT-4 and Claude 2.1 are employed for reasoning and managing smart home devices; MobileGPT~\cite{lee2024mobilegpt} on smartphones, which uses GPT-4 to manage mobile applications; and Intrusion Detection System Agent~\cite{li2024ids}, where GPT-4o reasons over network traffic data, generating and executing actions such as data preprocessing, classification, and knowledge retrieval for intrusion detection with detailed explanations.
In scenarios where a single LLM may exhibit limitations across diverse domains, a multi-LLM approach can be employed. This involves using multiple LLMs from different cloud providers, dispatched by a router. Upon receiving a user request, the router intelligently directs the query to the model best specialised for that specific domain, optimizing the response quality. For instance, Nexa AI's Octopus v4~\cite{chen2024octopusv4} is an LLM agent designed to run locally on the user's device. It analyzes incoming requests and determines the most appropriate model to handle them, preparing the prompt to maximise the chosen model's performance. This agent can route calls to both cloud-hosted and edge-hosted LLMs.
Additionally, Division-of-Thoughts (DoT)~\cite{shao2025division} is a framework that combines SLMs with powerful cloud LLMs to efficiently handle complex tasks. It first decomposes a user's query into simpler sub-tasks using a Task Decomposer, exploiting the reasoning abilities of language models. A Task Scheduler then analyzes dependencies among sub-tasks to decide which ones can be executed locally and which need cloud support. A lightweight, plug-and-play Adapter helps the SLM decide task allocation without changing its core parameters. This collaboration reduces costs, boosts speed, and preserves reasoning quality. 

\paragraph{Edge/Fog Distribution}
Edge and fog computing deploys models closer to users to minimise communication latency, while simultaneously maintaining significant computational power by distributing workloads across multiple devices. This approach enhances user privacy and system modularity, as models operate within user-managed nodes, ultimately improving overall system responsiveness.
Although lacking the computational power of cloud environments, these distributed architectures effectively harness the combined resources of numerous edge or fog nodes to distribute computation. This collaborative use of resources enables the deployment of larger, more capable models and the achievement of faster inference times compared to the constraints of a single edge device, as previously outlined. However, a significant challenge lies in the complexity of managing and maintaining the distributed nodes and orchestrating the computation across them, requiring considerable effort from the user.
The literature presents several methodologies to address this distributed deployment paradigm.
Some strategies focus on dynamic resource allocation and intelligent task distribution such as SpeziLLM~\cite{zagar2025dynamic} and AI Flow~\cite{shao2025ai}. SpeziLLM is an open-source framework that dynamically distributes LLM inference across decentralised fog and edge layers, in healthcare applications. By abstracting orchestration tasks like node selection, model placement, and task splitting, SpeziLLM simplifies the integration of LLMs into mobile and healthcare environments. It prioritizes the execution of sensitive data tasks on trusted local or fog nodes while offloading less critical or computationally intensive tasks to the cloud when necessary, balancing privacy, cost, and user experience through seamless model migration, fault tolerance, and flexible scaling. AI Flow redefines communication by focusing on "intelligence flow" rather than raw data transfer, adapting to dynamic network conditions to optimise inference across devices, edge nodes, and cloud servers. Instead of transmitting raw data, AI Flow sends only critical extracted features, significantly reducing communication overhead. It adaptively assigns portions of the inference task based on available computational resources, network bandwidth, and real-time conditions, ensuring low-latency responses and efficient model execution even in fluctuating environments.
Other strategies work on collaborative inference among edge devices such as Distributed Mixture-of-Agents (MoA)~\cite{mitra2024distributed}. This architecture enables multiple edge devices, each hosting a localised LLM, to collaborate through decentralised gossip protocols, achieving high-quality responses without a centralised server. Each device can independently process prompts and share intermediate results with neighboring devices using decentralized gossip algorithms. Devices act as "proposers" generating answers and "aggregators" refining or selecting the best response. This distributed setup enhances robustness, reduces latency, and improves answer quality compared to relying on a single device, while also ensuring queue stability despite resource limitations and varying workloads.
Another approach is using the edge devices to model partitioning and distribution, cooperating as a single high performance edge device. 
EdgeShard~\cite{zhang2024edgeshard} is a method that facilitates efficient LLM inference by partitioning large models into smaller "shards" and distributing them across multiple edge devices. It intelligently selects devices and allocates model parts based on their computing power, memory, and network conditions, leveraging collaborative edge computing to reduce latency, bandwidth usage, and privacy risks. By employing dynamic programming algorithms for optimized device selection and task scheduling, EdgeShard enables large models like Llama2-70B to run efficiently even in heterogeneous, resource-limited environments, significantly improving inference speed and throughput without compromising model accuracy.
In~\cite{zhao2024edge} the edges are used for a cooperative inference with terminal devices. In this framework efficient LLM inference is enabled by promoting collaboration between the user's device (terminal) and a nearby edge server. The terminal device rapidly generates speculative tokens using a lightweight model, while the edge server concurrently verifies and corrects them using a larger, more accurate LLM. This serial-parallel approach significantly reduces token generation delay and energy consumption by balancing the computational load between local and edge resources without heavy reliance on the cloud. An optimization algorithm manages model approximation and token generation to minimise delay and maintain high accuracy.
Finally in distributed scenarios, guaranteeing service despite intermittent connectivity is crucial. In~\cite{zimmer2024mixture} the authors address this problem with a novel "Mixture of Attentions" architecture for speculative decoding. This method significantly enhances a local small model's autonomous prediction by integrating Layer Self-Attention and Cross-Attention, effectively using LLM activations when available, yet maintaining accuracy when disconnected. This approach boosts robustness, enabling continuous, accurate inference without persistent network access.

\subsection{LLMs Alignment Strategies}
LLMs are pretrained on vast quantities of internet text, equipping them with remarkable capabilities across a spectrum of tasks, including conversation, mathematics, logic, reasoning, translation, and coding.
However, challenges emerge when LLMs, particularly within agentic frameworks, are tasked with operating in highly specialized or entirely novel domains, a common occurrence in pervasive environments.
For instance, deploying an agent to autonomously interact with smartphone applications, many unseen during the model's pretraining, often results in poor task completion performance.
Furthermore, the inherent resource constraints of pervasive computing often necessitate the use of smaller LLMs, which inevitably exhibit a notable performance decrease compared to their larger, more robust counterparts.
To address these limitations, a crucial technique is LLM alignment.
LLM alignment involves additional post-training of these models to specialise (align) them with the specific domains of deployment, thereby minimizing errors and hallucinations.
The primary method for achieving domain-specific alignment in pervasive computing is model fine-tuning, broadly categorized into three main approaches: Supervised Fine-Tuning (SFT), Direct Preference Optimization (DPO), and Federated Learning.

\paragraph{Supervised Fine Tuning}
Supervised Fine-Tuning aligns a LLM to a specific domain by training it on a curated reference dataset comprising input and desired output, $\mathcal{Z}=[x_{i}, y_{i}]$,  pairs relevant to the desired specialization. For instance, to align a model for mathematical tasks, the training data would consist of numerous examples of mathematical problems paired with their corresponding solutions.
Regarding dataset creation, three primary methodologies are employed:
\begin{itemize}
    \item Datasets can be constructed through human annotation of relevant examples.
    \item Alternatively, larger and more capable models can be used to generate task completions, with their interactions meticulously annotated until the desired outcome is achieved.
    \item Finally, another method involves human supervision of a large model, ensuring its adherence to the task and its correct progression towards the intended goal.
\end{itemize}
A common SFT approach involves full fine-tuning, where the model's initial weights, denoted as $\Phi_0$, are updated to $\Phi_0 + \Delta\Phi$ through iterative gradient descent. This process aims to maximise the conditional language modeling objective, as represented by:
\[
\max_{\Phi} \sum_{(x,y) \in \mathcal{Z}} \sum_{t=1}^{|y|} \log\left( P_{\Phi}(y_t \mid x, y_{<t}) \right)
\]
where $\mathcal{Z}$ represents the training dataset of (input x, target output y) pairs, and $y_{<t}$ denotes the preceding tokens in the target output sequence.

However, a significant drawback of full fine-tuning is that each downstream task necessitates learning a new set of parameters, $\Phi$, whose size is equivalent to the original model's parameter set, $\Phi_0$. In resource-limited scenarios, such as pervasive computing environments, this approach becomes prohibitively complex and computationally expensive.
To mitigate these challenges, Parameter-Efficient Fine-Tuning (PEFT) methods have emerged. These strategies are specifically designed to enable model adaptation even under stringent resource constraints.
Among prominent PEFT techniques are Low-Rank Adaptation~\cite{hu2022lora} (LoRA) and its quantized variant, QLoRA~\cite{dettmers2023qlora}. These methods enable efficient fine-tuning by learning a significantly smaller set of task-specific parameters, denoted as $\Theta$ , where $|\Theta| \ll |\Phi_0|$. Consequently, the task of determining the weight update $\Delta\Phi$ is reframed as an optimization problem over the smaller parameter set $\Theta$:
\[
\max_{\Theta} \sum_{(x,y) \in \mathcal{Z}} \sum_{t=1}^{|y|} \log\left( p_{\Phi_0 + \Delta\Phi(\Theta)}(y_t \mid x, y_{<t}) \right)
\]
This approach significantly reduces the memory and computational overhead associated with fine-tuning, making model adaptation more feasible in resource-constrained settings.
For instance InfiGUI Agent~\cite{liu2025infiguiagent} uses Full Supervised Fine-Tuning in two stages.
In Stage 1, they collected diverse page contents and vision-language datasets, using screen coordinates, to train basic skills like page contents understanding and grounding. In Stage 2, they synthesized new data to teach two advanced reasoning skills: hierarchical reasoning (strategic + tactical task planning) and expectation-reflection reasoning (self-correction from outcomes).
For dataset creation, they used both real page contents datasets and synthetic SFT data generated from multimodal LLMs. They focused on "Reference-Augmented Annotations" to precisely link visual elements and textual reasoning.
ReachAgent~\cite{wureachagent} uses a first stage of Supervised Fine-Tuning (SFT).
They build three datasets: Page Navigation, Page Reaching, and Page Operation. Each dataset focuses on different subtasks such as reaching pages, operating within pages, and navigating multi-step tasks. They generated tasks and step-by-step labels using various LLMs. The SFT trains the model to understand full page contents flows and improve multi-step planning before any reinforcement learning. This stage ensures the agent can solve subtasks accurately before optimizing for full-task preferences in later RL training.
MaViLa’s Supervised Fine-Tuning (SFT) to perform visual scene understanding, anomaly detection, and manufacturing reasoning using using LoRA on a Vicuna 13B model~\cite{fan2025mavila}. For dataset creation, they collected real-world and schematic manufacturing images, each manually captioned.
They generated instruction-response pairs by prompting GPT-4, distinguishing between general and domain-specific questions. For domain-specific instructions, they used Retrieval-Augmented Generation (RAG) to ground answers in manufacturing knowledge. Instructions were classified by complexity, reasoning need, and domain specificity to ensure high-quality fine-tuning data.
Finally in~\cite{yonekura2024generating} researchers  used Supervised Fine-Tuning (SFT) to customise LLMs for stable activity generation in smart home simulations.
To create the fine-tuning dataset, they collected labeled examples of human-like daily schedules and activity outputs.
Fine-tuning focused on ensuring structured outputs (like JSON) to prevent simulator crashes from unstructured LLM replies.
The SFT enhanced the model’s ability to generate realistic, context-aware daily activities for virtual smart home agents. As a result, they achieved a 4.3\% improvement in simulation stability and efficiency compared to baseline prompting.

\paragraph{Direct Preference Optimization}
Another effective approach for aligning LLMs through Fine-Tuning is Direct Preference Optimization (DPO)~\cite{rafailov2023direct}. This method offers a more efficient approach to instruction-tuning. It directly trains the LLM agent on pairs of preferred ("winner") and less-preferred ("loser") responses to an instruction. Critically, DPO achieves this by fine-tuning LLMs without the need for an explicit reward model, relying solely on these direct preference comparisons between output pairs.
The DPO algorithm begins by sampling completions $y_1, y_2 \sim \pi_{\text{ref}}(\cdot \mid x)$  
for each prompt $x$, and then labels these completions based on human preferences to construct an offline preference dataset: 
\[
\mathcal{D} = \left\{ \left(x^{(i)}, y_w^{(i)}, y_l^{(i)} \right) \right\}_{i=1}^{N}
\]
Here, $x^{(i)}$ represents the $i_{th}$ input prompt, $y_w^{(i)}$ denotes the preferred ("winner") output, and $y_l^{(i)}$ represents the "loser" output, as determined by human feedback.
Then, the language model $\pi_\theta$ is optimized by minimizing the DPO loss:
\[
\mathcal{L}_{\text{DPO}}(\pi_\theta; \pi_{\text{SFT}}) = 
- \mathbb{E}_{(x, y_w, y_l) \sim \mathcal{D}} \left[
\log \sigma \left(
\beta \log \frac{\pi_\theta(y_w \mid x)}{\pi_{\text{SFT}}(y_l \mid x)} - 
\beta \log \frac{\pi_\theta(y_w \mid x)}{\pi_{\text{SFT}}(y_l \mid x)}
\right)
\right]
\]
Since these datasets are typically sampled using a Supervised Fine-Tuned model $\pi_{\text{SFT}}$, the reference policy is often initialized as $\pi_{\text{ref}} = \pi_{\text{SFT}}$ when available.
However, if $\pi_{\text{SFT}}$ is not accessible, the initialization of $\pi_{\text{ref}}$ is achieved by maximizing the likelihood of the preferred completions $(x, y_w)$ within the preference dataset:
\[
\pi_{\text{ref}} = \arg\max_{\pi} \, \mathbb{E}_{(x, y_w) \sim \mathcal{D}} \left[ \log \pi(y_w \mid x) \right]
\]
For instance ReachAgent~\cite{wureachagent} uses Direct Preference Optimization (DPO) to refine its decision-making in mobile GUI tasks. DPO is used without needing explicit numeric rewards, instead relying on preference pairs indicating which actions are better. These preferences are constructed using a 4-level reward ranking (Golden > Longer > Incomplete > Invalid) based on how effectively and efficiently page contents flows complete the task. During training, the model learns to prefer actions that lead to more optimal flows—those that are both task-completing and concise. 
This preference data is fed into the DPO loss function, which guides the policy ($\pi_\theta$) to align closer to preferred behaviors while softly deviating from the supervised fine-tuned policy ($\pi_{SFT}$). This enhances the model’s ability to generate efficient and successful page contents flows without needing exact matches to gold actions.

\paragraph{Federated Learning}
Federated Learning (FL) is a machine learning paradigm that enables the training of a unified model across numerous decentralized edge devices or servers, each holding local data, without the need to exchange these sensitive datasets~\cite{bimpas2024leveraging}. Essentially, FL allows for collaborative algorithm training on distributed data sources while preserving data locality. One of the core principle of FL is to empower devices, including smartphones and IoT devices, to collectively learn a shared predictive model, distributing the training computation between multiple devices.
For instance in FedMobileAgent framework~\cite{wang2025fedmobileagent}, federated learning was used to collaboratively train mobile agents across decentralized user data while preserving privacy.
First each user locally collects data $D_k = \left\{ \langle T, a_i, s_i \rangle \right\}_{i=1}^n$ via Auto-Annotation, where $T$ is a task instruction, $a_i$ an action, and $s_i$ a screenshot.
Then local training updates the model using stochastic gradient descent:
\[
M_k^{(l, r+1)} = M_k^{(l, r)} - \eta \nabla \ell\left(M_k^{(l, \tau_k)}; T, s, a\right)
\hspace{0.5em}
\text{where } M_k^{(l, \tau_k)} \text{ is sent to the server, } 
M^{(l+1)} = \sum_{k \in S^l} \omega_k M_k^{(l)}, 
\text{ and } \omega_k = \frac{n_k^*}{\sum_{k \in S^l} n_k^*}.
\]
This two-level weighted aggregation balances episode and step diversity, improving learning from non-Independent and Identically Distributed data.

\subsection{Agent Design Strategies}
As discussed in Section~\ref{sec:agents}, LLM-based agents use Large Language Models to perceive their environment, reason and plan into actionable subtasks, execute actions for each subtask, and iteratively refine their approach based on feedback. Furthermore, operating within a pervasive environment, often characterized by resource-constrained devices or distributed systems such as cloud, fog, and edge nodes, introduces communication overhead and latency. Consequently, the design of an agent's architecture requires careful consideration not only of the underlying LLM deployment strategy but also of Agent module adaptation, which includes Memory, Reasoning, Planning and Action modules, as these significantly impact the agent's behaviour and overall performance within these constrained settings.

\subsubsection{Memory}
\label{sec:memoryperv}
Memory is a critical component of intelligent agents, enriching their perception of the environment by providing crucial contextual information. It enables the agent to not only understand the immediate state of the external world but also access to historical data, including past execution trajectories, actions previously undertaken, errors encountered in prior attempts, and potentially effective solutions that can enhance future performance.
In pervasive computing agents typically implement two primary forms of memory: Short-Term Memory and Long-Term Memory.

\paragraph{Short-Term Memory} This memory saves the intermediate steps within its current execution cycle. These ongoing steps are readily accessible at each decision-making stage, allowing the agent to reason about the next action in a context-aware manner. This mechanism significantly reduces the likelihood of the agent becoming trapped in repetitive states or endlessly cycling through the same sequence of actions. For instance in~\cite{wang2024mobile} the agent employs short-term memory to meticulously record all actions performed and the corresponding state changes as it interacts with the environment. This dynamically observed information then directly informs the next operational decisions. Also in~\cite{li2024ids} the agent uses short-term memory to maintain a log of the current session's context, including all prior reasoning steps, executed actions, and received observations. This memory is structured and iteratively updated after each tool execution, ensuring a coherent and up-to-date understanding of the ongoing situation. This allows the LLM to generate contextually relevant thoughts and actions at every stage of its processing pipeline. Crucially, this short-term memory is session-bound and discarded upon the completion of the inference process, guaranteeing real-time decision traceability without necessitating long-term storage of transient data.

\paragraph{Long-Term Memory} Once an agent concludes its execution, regardless of whether the final goal was achieved, all the intermediate steps and experiences accumulated during the session are consolidated. Through various summarization or simplification techniques, this information is then archived in Long-Term Memory for future reference and learning.
For instance in~\cite{rivkin2024aiot} the agent leverages long-term memory to store a comprehensive history of user interactions with the agent. It employs a vector database, using the MiniLM embedding model for efficient storage and retrieval of semantically similar past executions when confronted with new tasks. Additionally, high-level summaries of user-agent interactions are stored to build a dynamic and holistic understanding of individual user preferences over time.
Mobile-Agent-E~\cite{wang2025mobile} manages memory through a persistent long-term storage, focusing on two key elements derived from past experiences: Tips and Shortcuts. Tips represent generalizable insights gleaned from prior tasks, providing guidance for both high-level strategic planning and low-level action execution. Shortcuts are reusable sequences of actions identified for frequently occurring subroutines. Following each completed task, two "Experience Reflectors" analyse the entire interaction history to update existing Tips and Shortcuts or generate new ones. These refined or novel insights are then using by the "Manager" for future planning and the "Operator" for the next action execution. Finally in~\cite{lee2024mobilegpt}, the agent employs a hierarchical memory system that stores tasks as ordered sequences of subtasks and actions, directly linked to specific application screens. Each screen is represented as a node containing a set of available subtasks, with edges denoting transitions between subtasks triggered by related low-level actions. Tasks are saved in a parameterized function-call format, enabling flexible reuse across different contexts. During execution, the system can recall previously encountered tasks or subtasks and adapt them to new situations through attribute matching and in-context learning techniques. The memory is dynamically updated based on user feedback or the agent's own self-correction mechanisms. 

As previously discussed, long-term memory serves as a repository of both successful and failed executions, enabling the agent to learn and improve over time. Two core strategies are commonly employed for populating this valuable resource:
\begin{itemize}
    \item \textbf{Exploratory Phase:} Often, before being deployed for user interaction, an agent goes through a dedicated exploratory phase. During this stage, the agent is encouraged to actively explore its environment (e.g., the functionalities of various applications on a device). The primary objective is to develop a foundational understanding of the interaction mechanisms, identify the available tools and their functionalities, and learn the effects of different actions and strategies for overcoming specific challenges. This proactive exploration ensures that, when the agent is eventually used for real tasks, it already possesses a significant knowledge to more reliably navigate the environment and find the correct path to achieve its goals. For instance MobileGPT~\cite{lee2024mobilegpt} analyzes app screens offline, extracting UI layout and simplifying it to HTML. The LLM identifies subtasks, formats them as function calls, and caches them for efficient live execution. AutoDroid~\cite{wen2024autodroidv1} explores apps by random UI interactions, building a UI Transition Graph summarizing states and elements as simulated tasks, stored in memory. Also AutoDroid-V2~\cite{wen2024autodroid} constructs structured documents from page contents traces, abstracting states and transitions.
    \item \textbf{Test Phase:} In this scenario, the agent continuously learns from its interactions with the user during live testing. Both successful and failed execution attempts are stored in long-term memory. This continuous learning process allows the agent to gradually refine its strategies and improve its performance as the user engages in more tasks. For instance IDS Agent~\cite{li2024ids} uses long-term memory to resolve ambiguous situations by retrieving past sessions similar to the current one, incorporating their reasoning and outcomes for better decisions. Also the LiMeDa Framework~\cite{chen2024llm} stores summarized data from completed vehicle tasks (route, time, energy) in memory, managing it to prevent overload. Upon receiving a new task, LiMeDa retrieves relevant past experiences to improve decision efficiency and avoid repeating errors.

\end{itemize}

\subsubsection{Reasoning and Planning}
Following the perception of the system state and the memory module's contents, including previous actions or complete past executions, the agent proceeds through Reasoning and Planning.
The Reasoning step bring out a logical thinking process from the LLM, facilitating the formulation of an effective plan.
For instance Mobile Agent-E~\cite{wang2025mobile} implements an Action-Reflection module that analyzes the state before and after an action, along with the action itself, to determine if the outcome aligns with the expected goal. This module categorizes action outcomes into: Successful or partially successful (outcome matches expectation), Failed because result leads to an incorrect state, and Failed because an action produces no observable change.
Also InfiGUI Agent~\cite{liu2025infiguiagent} operates in a three-step cycle: Reasoning (performing hierarchical reasoning), Action generation (writing the next action and its anticipated outcomes), and Reflection (analyzing the resulting state to evaluate if the expected results were achieved and generating a textual summary of this reflection).
A popular reasoning technique is Chain of Thought (CoT) prompting~\cite{wei2022chain} , that allows the agent to tackle complex reasoning tasks by breaking them down into a sequence of explicit steps, often initiated by prompts such as "Let's think step by step".
For instance Mavila~\cite{fan2025mavila} , used for process automation, employs CoT reasoning to decompose tasks into a series of sequential steps and in the SAGE framework~\cite{rivkin2024aiot}, planning (the decomposition of a high-level goal into substeps) is managed using CoT reasoning in conjunction with the ReAct~\cite{yao2023react} design pattern. Tool instructions and formatting guidelines encourage the LLM to first outline a plan before specifying the execution details. Also AutoDroid~\cite{wen2024autodroidv1} fine-tunes a small language model (SLM) to reason with a zero-shot CoT approach, using a structured format.
\newline
The other step is Planning, where, informed by prior events and the reasoning process, the agent decides on the next action(s) to take. For instance in the IDS Agent~\cite{li2024ids}, planning involves selecting the appropriate tool based on the information in memory and the analyzed traffic data. The agent then decides, based on classification results, whether to trigger an alert, thus identifying a potential security threat. Mobile Agent-E~\cite{wang2025mobile} features a Manager module responsible for generating a plan broken down into subtasks. At each step, the Manager considers the initial user query, the current screenshot, the previous overall plan, the preceding subgoal, the current progress status, available Shortcuts from long-term memory, and any relevant notes. It then updates the overall plan and identifies the next immediate subgoal to pursue. Finally in~\cite{wang2024mobile} the agent employs three distinct agent roles for mobile device interaction. The Planning Agent determines the task progression using historical data. The Decision Agent observes relevant content from past screens via memory to generate actions and update the memory. The Reflection Agent evaluates if actions meet expectations by comparing screen states and initiates corrective re-execution if needed.

\subsubsection{Action}
The specific actions that an agent can perform within pervasive computing are highly dependent on its designated operational domain and the available tools or interfaces. Common categories of agent actions include:
\begin{itemize}
    \item \textbf{Tool Use / API Calls:} Agents can leverage external services or system functionalities by using specific tools or making API calls. This includes a wide range of interactions, such as operating smartphone applications, sending emails, querying databases, retrieving weather information, or controlling external devices. For example, in~\cite{wen2024autodroidv1}, the agents possesses a defined set of possible actions (CLICK, CHECK/UNCHECK, SCROLL<DIRECTION>, INPUT<TEXT>) to interact with a smartphone device's interface. In~\cite{rivkin2024aiot}, the agent flexibly manages smart home devices using a dedicated device interaction tool. The API calls are executed using the SmartThings REST API for reading device attributes and sending commands, with an automatic error correction mechanism in case of call failures. Similarly, in~\cite{li2024ids} the agent is equipped with a series of tools, including Knowledge Retrieval, Data Extraction, Classification, and Long-Term Memory Retrieval, which it can use for intrusion detection analysis. These tools are invoked using a JSON format that specifies an action name (the tool's identifier) and an action input (the associated settings or parameters for that tool).
    \item \textbf{Code Execution:} Agents may generate and execute code snippets to test conditions or directly interact with underlying systems. For example, in~\cite{wen2024autodroid}, the agent interacts with the environment by translating a user's natural language task into executable Python-like scripts. These scripts are generated by a small language model based on a structured app document and are then executed on the device to manipulate page contents elements, such as tapping buttons or scrolling content. Also, in~\cite{rao2024eco}, a large language model is employed to generate Python code that guides the placement of microservices between edge and cloud infrastructure, based on real-time workload and latency data. This enables the LLM to adaptively recommend whether a microservice should run on the edge or in the cloud, with new placement decisions being automatically integrated into the running application.
    \item  \textbf{User Messaging (Natural Language Interaction):} Agents can also act by directly interacting with users through natural language messages, providing support as intelligent assistants. For instance, in~\cite{fan2025mavila}, the agent communicates its analysis of manufacturing images to users, performing anomaly detection and scene understanding to provide comprehensive support to smart factory employees. Also, in~\cite{zagar2025dynamic}, the system uses a unified interface to interact with users naturally via chat, enabling secure, real-time processing of health data in healthcare domains such as Electronic Health Record (EHR) analysis, patient data explanation, and medical form automation.
\end{itemize}

\subsection{Evaluations Metrics}
The evaluation of LLM-based agents in pervasive computing environments must consider various variables such as the strict constraints of edge hardware or user personalization. 
\paragraph{Resource consumption} Must be quantified to evaluate the cost of usage of an LLM Agent in resource contraint devices. Energy consumption \cite{zhao2024edge} is a critical metric for battery-powered devices. It measures the Joules expended per inference task or per token generated. Studies on mobile agents often track the Battery Drain Rate (\%/hour) during active agent operation compared to idle states. Also some measure track the inference latency such as Time Per Output Token (TPOT) for generative tasks \cite{shao2025ai}. For shared services, Context Switching Latency measures the overhead of loading different user profiles or agent personas into active memory on edge nodes \cite{yin2024llm}. Finally Memory Footprint is used to track the RAM usage of the agent, specifically the Peak Memory Usage during inference and the KV Cache Size \cite{yin2024llm}. 

\paragraph{Contextual Accuracy and Personalization} Agents in pervasive environments must bridge the gap between ambiguous sensory inputs and specific user needs, requiring them to infer intent and adapt to personal preferences in real-time. In this case Implicit Preference Accuracy measures the agent's ability to predict the correct item recommendations. Element Accuracy (Ele.Acc) measures the correctness of predicted UI coordinates on mobile screens, while Intersection over Union (IoU) evaluates the precision of bounding boxes when the agent must visually identify IoT devices or physical objects \cite{zhang2024smartagent}. Finally Intent Resolution \cite{rivkin2024aiot} measures the agent's success in disambiguating vague voice commands (e.g., "make it cozy") into specific device actions (e.g., "dim lights to 50\%, set temperature to 22°C").

\subsection{Applications of Pervasive Agents}
As discussed in Section~\ref{sec:pervasive}, while pervasive computing offers a multitude of applications, it still faces with challenges such as extracting and integrating heterogeneous data from diverse sources, translating complex natural language instructions into standardised actions executable across various tools and devices, and navigating intricate environments to accomplish complex objectives.
These challenges can be effectively addressed by LLM-agents. As detailed in Section~\ref{sec:pervasiveagent}, these systems possess a suite of features capable of overcoming these limitations and also unlocking a broad range of potential applications within the pervasive computing domain.
\paragraph{\textbf{Smart Homes}} Agents interpret user commands for device control via dynamic planning. For instance  SAGE agent achieve 76\% success rate on complex queries~\cite{rivkin2024aiot}. Multi-agent systems like CASIT enhance IoT deployments by coordinating sensor data and detecting anomalies, outperforming single agents in data-rich environments~\cite{zhong2024casit}. The MuRAL dataset aids socially aware LLM development for smart environments with richly annotated sensor data~\cite{chen2025mural}. Frameworks like LLMind integrate LLMs with AI modules for multi-device orchestration via natural language commands translated into device control scripts~\cite{cui2024llmind}.

\paragraph{\textbf{Mobile Task Automation}} Agents automate smartphone tasks by interpreting user instructions and executing page contents actions, eliminating manual interaction. AutoDroid~\cite{wen2024autodroidv1} use UI understanding and exploration for autonomous task completion. InfiGUIAgent~\cite{liu2025infiguiagent} employs human-like reasoning for robust UI interaction. MobileGPT~\cite{lee2024mobilegpt} prioritizes efficiency with adaptable task memory and hybrid failure recovery. Benchmarks like Android Agent Arena (A3)~\cite{chai2025a3}, LlamaTouch~\cite{zhang2024llamatouch}, and MobileAgentBench~\cite{wang2024mobileagentbench} facilitate agent evaluation, while MobileSafetyBench~\cite{lee2024mobilegpt} focuses on safe handling of sensitive operations.

\paragraph{\textbf{Smart Health and Factories}} Agents provide analysis of medical and industrial data. AutoHealth~\cite{cardenas2024autohealth} integrates agents into wearables for Parkinson's monitoring. SpeziLLM~\cite{zagar2025dynamic} enables privacy-preserving medical AI on local nodes. In Smart Manufacturing, IMVA~\cite{lin2025generative} and MaViLa~\cite{fan2025mavila} offer multimodal reasoning for tasks like quality control and system orchestration, improving decision-making and automation.

\paragraph{\textbf{Smart Cities}} Agents enable intelligent vehicle dispatching (LiMeDa)~\cite{chen2024llm}, realistic personal mobility generation (LLMob)~\cite{jiawei2024large}, and natural language interaction with smart building management for energy optimization and control (BuildingSage)~\cite{dedeoglu2024buildingsage}. These systems demonstrate the transformative potential of LLM-based agents in creating intelligent and adaptive urban infrastructures.

\section{Discussion}
LLM-based agents represent a substantial advancement in Artificial Intelligence, enabling the tackling of highly complex tasks across diverse domains including software engineering, research, medicine, and robotics. As outlined in Section~\ref{sec:agents}, these agents operate by perceiving their environment, reasoning and planning actions to achieve defined goals, and executing these plans through specific actions. This architecture typically uses a foundational model, either an LLM or a MLLM, and three essential modules: Memory, Planning, and Action. The Memory module stores necessary information, the Planning module designs action sequences, and the Action module generates and applies individual steps.
Pervasive computing stands out as a field for a practical application of this architecture. As detailed in Section~\ref{sec:pervasive}, this domain focuses on enhancing human activity through integrated technology, providing computational power for a variety of applications that can significantly improve daily life. From smart homes and factories to the every day usage smartphones, embedding the computational power of agents within pervasive computing systems holds the potential to solve numerous challenges by autonomously handling previously intractable tasks. Section~\ref{sec:pervasiveagent} introduced various strategies for integrating these agents into pervasive devices. These strategies are heavily influenced by the available resources on the deployment devices, and have facilitated the application of agents in pervasive computing, including personal assistants for health monitoring and device interaction, as well as building and traffic management in smart cities.
However, several critical challenges remain under active research and warrant careful consideration.
\paragraph{Small and Quantized Models Limitations} A primary challenge in pervasive computing is the resource intensity of LLMs relative to edge hardware constraints. As discussed in Section~\ref{sec:deploym}, deployment often necessitates the use of Small Language Models (SLMs) or aggressive quantization to meet energy and privacy requirements. However, a critical research gap specific to agentic AI is the degradation of reasoning action capabilities. When models have less parameters, their ability to maintain long-horizon goals, adhere to complex tool-use syntax, or self-correct during multi-step planning fails significantly more often than larger counterparts. Furthermore, smaller models are more prone to hallucinations, generating plausible but factually incorrect token sequences, which can lead to catastrophic failure in autonomous decision-making loops.

To address these limitations, future research must move beyond general knowledge distillation. The field requires specialized reasoning distillation, where SLMs are trained specifically to retain agentic behaviors (e.g., planning logic, API error handling). Additionally, research should focus on dynamic compute allocation frameworks that can assess task complexity in real-time, solving easy questions locally while dynamically offloading reasoning-heavy questions to larger remote models only when necessary. Finally, optimizing retrieval-augmented generation (RAG) specifically for low-parameter models is crucial to mitigate hallucinations without increasing model size.

\paragraph{Memory Limitation}
The external nature of the Memory module in current agent architectures also presents a significant limitation. As agents interact with their environment, their experiences must be stored externally and then selectively reintroduced into the LLM's context window. The limited size of this context window poses a constraint, particularly for agents requiring numerous interactions to achieve a goal, potentially leading to the loss of crucial information if earlier interactions are simplified or overly summarized, as discussed in Sections~\ref{sec:memory} and~\ref{sec:memoryperv}.
To overcome this, researchers are exploring strategies to find better summarization methods, or to expand the context window, as seen with the 1 Million context window Qwen's Qwen2.5-Turbo~\cite{yang2025qwen2} and Google's Gemini 2.5 Pro. Another promising direction involves the direct integration of memory modules within the models themselves. For instance, in~\cite{behrouz2024titans}, the Google researchers propose a novel architecture incorporating a neural long-term memory module. This module learns to memorize historical context at test time, demonstrating superior effectiveness over both Transformers and recent linear recurrent models, particularly in long context scenarios.
Research should therefore concentrate on optimizing these on-device neural memory techniques and developing advanced context summarization algorithms that can maintain an effective agent execution trajectory without rapidly consuming the available context window.

\paragraph{Energetic Issue}
The Energetic issue is particularly significant when we are facing with agents that have to use frequently LLMs for reasoning, planning, action, and memory management tasks. This challenge must be addressed especially in pervasive environments, where there is a wide heterogeneity of devices, from large cloud high resources computing servers to smartphones that have limited energy resources.
As discussed in Section~\ref{sec:deploym}, an LLM has a very high energy cost per billion parameters, which leads to rapid battery drain on edge devices and high electricity consumption when used on fog or cloud servers.
The research challenge is how to achieve autonomous function while minimizing the energy cost of the continuous perceive-plan-act loop.

To mitigate this, future research must focus on optimizing both the core inference process and the agent's decision cycle. Techniques that reduce computation for every inference step are vital: this includes optimizing internal structures through methods like KV cache compression and swapping to avoid expensive recomputations~\cite{yin2024llm}, and adopting optimized LLM sharding strategies coupled with hardware techniques like power capping to improve energy per task~\cite{samsi2023words}. Furthermore, research must target the agent's decision-making process itself. This involves developing dynamic early-exiting inference frameworks like GREEN-CODE~\cite{ilager2025green}, which use reinforcement learning to terminate inference when sufficient confidence is obtained, achieving substantial energy reductions. Another key direction is implementing event-triggered planning architectures that use low-power sensors to detect semantic events and only activate the high-power LLM reasoning engine when a proactive, high-value action is necessary. This prevents continuous, costly reasoning for trivial or irrelevant environmental changes.

\paragraph{Privacy Issue}
Finally, in pervasive computing, it's fundamental to prioritise privacy because of its close interaction with users and the resulting handling of large amounts of sensitive data. Agents operating in this context must manage this information while guaranteeing the user's complete privacy.
For instance, when an agent interacts with applications on a smartphone or PC, it must exercise extreme caution in handling sensitive user data, including full names and the contents of private documents or files. The literature features various analyses and several benchmarks specifically designed to evaluate the security of agents in ensuring the preservation of user privacy.
The PrivacyLens framework~\cite{shao2024privacylens} provides a multi-level evaluation of LLM agents’ awareness of contextual privacy norms. It introduces a pipeline that converts privacy-sensitive seeds into expressive vignettes and executable agent trajectories to uncover instances of private information leakage, such as sharing job-seeking details inappropriately, even with privacy-preserving prompts. PrivacyLens reveals that even advanced LLMs like GPT-4 leak sensitive information in a significant number of cases (up to 25.68\%).
AgentDojo~\cite{debenedetti2024agentdojo} assesses agents robustness against prompt injection attacks. This framework puts LLM agents with realistic scenarios like email management or banking app navigation while defending against malicious inputs aimed at extracting user data. AgentDojo highlights the persistent vulnerability of agents to adversarial inputs and underscores the need for privacy-aware defenses.
Finally, Agent-SafetyBench~\cite{zhang2024agent} offers a broader evaluation of agents' behavior across diverse environments, identifying privacy-related safety risks like unintentional data leakage. Evaluation across 2,000 test cases revealed that no agent achieved a safety score above 60\%, highlighting a systemic lack of robustness and risk awareness in current LLM-based agent implementations.
\newline
While current mitigation strategies like Personal Identifiable Information (PII) scanner (e.g., AutoDroid~\cite{wen2024autodroidv1}) and layered firewalls~\cite{abdelnabi2025firewalls} have shown promise in reducing leakage rates, they often rely on static rules that can be avoided. Future research must evolve towards action-aware semantic firewalls, that do not merely filter text but analyse the intent and potential consequences of tool calls. For example, a firewall should dynamically restrict high-stakes API calls based on geolocation or user presence. Furthermore, research is needed into cryptographic verification methods that ensure critical physical actions originate from a verified human intent rather than an autonomous agent hallucination, creating a hard "human-in-the-loop" requirement for high-risk interventions.

\begin{figure}[t!]
    \centering
    \includegraphics[width=0.7\textwidth]{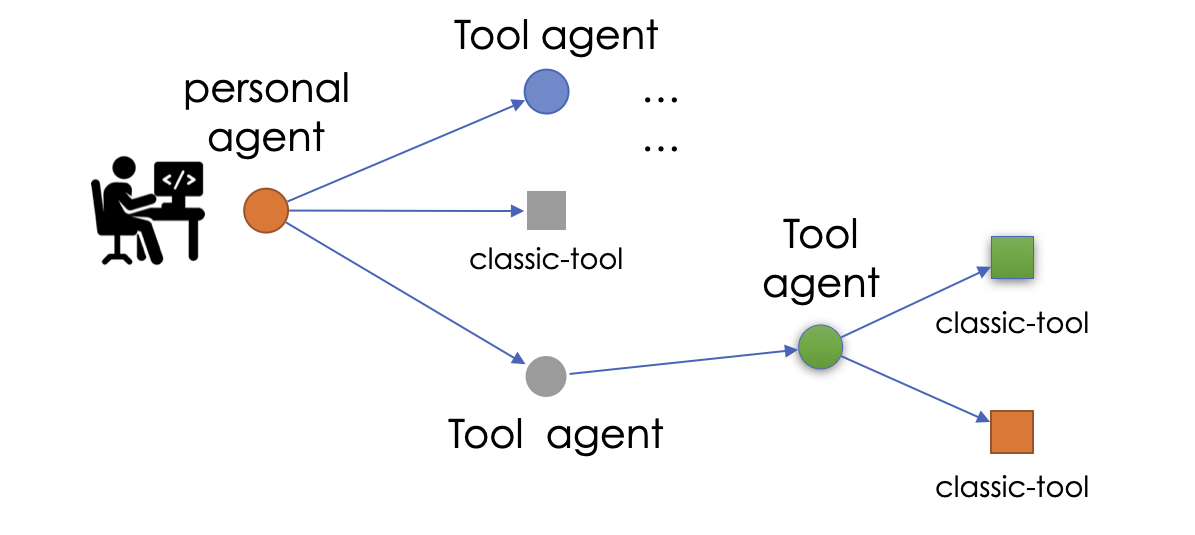}
    \caption{A topology of classic tools and agent as a tool}
    \label{fig:agentasatool}
\end{figure}
\section{Conclusion}
This paper investigates the evolving landscape of research on LLM-based agents in the domain of pervasive computing. Initially, it provides a comprehensive analysis of agents, exploring their architecture, evaluation methodologies, and diverse applications across various fields.
Afterwards, the paper introduces the pervasive computing field, outlining its infrastructures and wide-ranging applications. It then explores the growing integration of Gen-AI within this domain, culminating in the integration of intelligent agents.
Building on this foundation, the paper scrutinises agent-based architectures specifically designed for pervasive environments. This includes an analysis of architectural adaptations, novel strategies introduced to suit the constraints of these contexts, and a review of implemented applications.
The paper then discusses the critical challenges in this field, alongside an overview of current research efforts aimed to address these limitations.
\newline
The field of artificial intelligence, is experiencing a period of exponential growth. Each month brings the release of more robust and effective models, driven by novel LLM architectures such as Mixture of Experts~\cite{jiang2024mixtral} and increasingly efficient post-training alignment techniques such as reinforcement learning techniques like GRPO~\cite{shao2024deepseekmath}, knowledge distillation, and fine-tuning methods like SFT and DPO.
New powerful open-weight model families (e.g., DeepSeek r1, LLaMA 4, Qwen 3) and close-sources (e.g. Claude 4, OpenAI o1, o3) are released every few months. We are also observing the emergence of increasingly compact models that shows similar performances or even surpass larger state-of-the-art counterparts. For example, QwQ-32B demonstrates performance comparable to or better than models like DeepSeek-R1-Distilled-Qwen-32B, DeepSeek-R1-Distilled-LLaMA-70B, o1-mini, and the original DeepSeek-R1~\cite{QwenTeam_2025}. This miniaturization is critical for pervasive deployments.
\newline
Substantial progress is also being made in hardware development. Innovations like the cost-effective NVIDIA Jetson Orin and the NVIDIA DGX Spark project, featuring compact devices capable of running models with up to 200 billion parameters, clearly signal a future where affordable, personalised agent hosting is accessible to individuals and companies~\cite{NVIDIA_DGX_Spark_2025}. Also the NVIDIA \textit{CEO} Jensen Huang highlights this rapid advancement, stating, "\textit{Our systems are progressing way faster than Moore’s Law}", attributing this acceleration to integrated innovation across the entire stack of architecture, chip, systems, libraries, and algorithms~\cite{Zeff_2025}. This means that increasingly compact devices will offer greater computational power at lower costs. 
\newline
Over the next 2–3 years, the exponential growth in both of these areas will profoundly impact the agent field in pervasive computing, driven by the increasing performance of Edge Computing devices. The continuous integration of greater computational capabilities into smaller, more energy-efficient, and affordable hardware will enable the deployment of agents of various scales, dynamically adapting to user service requirements. Consequently, we will see a proliferation of specialized agents distributed across  edge nodes, each providing tailored services based on their computational resources and the robustness of their embedded LLMs.
This trajectory marks a remarkable shift in pervasive computing: from an "\textit{anytime, anywhere}" model to one that is "\textit{all the time, everywhere}".
The traditional "\textit{anytime, anywhere}" agent approach relied on a limited number of high-performance computing (HPC) devices providing agent services, which were constrained by connectivity and response latency. In contrast, the emerging "\textit{all the time, everywhere}" agent approach leverages the increasing capability of smaller devices, such as fog and edge nodes, to host agents. This ensures the continuous provision of agent services, virtually at all times. As detailed in Section \ref{sec:deploym}, various strategies for deploying LLMs, the core of these agents, have been proposed, as seen in works like~\cite{shao2025ai, zhao2024edge, zhang2024edgeshard, mitra2024distributed}, and research in this area is rapidly advancing.
This future will be characterised by an increasingly interconnected network topology spanning cloud, fog, and edge devices. The enhanced performance of these edge components will enable the robust deployment of LLM-based agent services much closer to the user, facilitating faster response times and ensuring continuity of service even under unstable connectivity. Moreover, this distributed computational capacity is the key enabler for Multi-Agent Systems, where specialized agents can collaborate across the edge-to-cloud continuum to solve complex tasks that would overwhelm a single isolated model. Crucially, a smaller, edge-deployed agents can maintain essential functionality and deliver ongoing service in scenarios where reliance on cloud resources would be impractical or impossible. This vision lead to an increasing proliferation and usage of various agent services in multiple domains and applications, raising a crucial question: 
\newline
\textbf{Does the future of pervasive computing necessitate single, general-purpose agents capable of autonomously operating across numerous domains, or will it favour multiple, specialized agents, each expertly designed to solve a dedicated task upon request?}
\newline
Consider a single agent attempting to manage both smartphone applications and smart home sensors. Achieving such broad expertise would demand computationally intensive, highly robust models and significant resources (a particular challenge in pervasive computing).
Conversely, the demonstrated effectiveness of domain-specific fine-tuning suggests a more practical and scalable approach: deploying separate, specialised agents for distinct tasks. This leads to what we call "\textbf{Agent as a Tool}", where each specialized agent offers as-a-service points of provision (Figure~\ref{fig:agentasatool}).
In a realistic future scenario, each individuals can have a personal agent on their devices. These personal agents will be trained to interact with classic tools (e.g., email, calendar)~\cite{wu2024avatar, liu2024apigen} and, critically, call other specialised agent-tools. For instance, a personal agent might seamlessly call upon a dedicated smart home agent for managing household systems or a software engineering agent for coding assistance within a development environment. The personal agent would intelligently decide which agent-tool to use for specific subtasks. Its overall task-handling capacity would depend on factors such as the robustness of its core LLM, the computational hardware constraints (edge, fog, cloud), and the accessibility of required classic-tools or agent-tools (e.g., network availability, physical location, co-hosting status). A calendar application, for example, might be locally accessible, while specialised company agent-tools might reside on remote edge or fog devices.
This vision culminates in a complex, compound system comprising personal agents, classic tools, and specialised agent-tools. While these agents will operate semi-autonomously, proposing plans that align with the specialized task's workflow and executing corresponding actions, user interaction and guidance will remain crucial. To enhance agent performance and user experience, advanced memory strategies, such as summarizing past interactions, can be employed. This interaction data also presents a valuable opportunity for further fine-tuning or reinforcement learning techniques to improve agent performance~\cite{wu2025collabllm}.
Another paramount aspect is security. For critical actions (e.g., database modifications, code commits), essential safeguards requiring authorized user approval are indispensable for both personal agents and agent-tools~\cite{patil2024goex, abdelnabi2025firewalls}. Robust firewall strategies must also be in place to protect sensitive user data. Communication between personal agents, tools, and agent-tools can leverage various established protocols, such as A2A\footnote{\href{https://developers.googleblog.blog/2024/05/a2a-a-new-era-of-agent-interoperability/}{A2A, Agent Interoperability}} for agent-to-agent communication and the Model Context Protocol (MCP)~\cite{unknown-author-2024} for classic tool calls.
This architecture's broad and inherent adaptability, applicable from optimizing industrial production to enhancing individual daily life, provides a significant opportunity to fundamentally reshape how businesses operate and to tangibly improve the daily experiences of people.

\begin{acks}
Funders: Fabio Ciravegna has received funding from the European Union’s HORIZON programme, project  “ICOS: Towards a functional continuum operating system”, grant agreement No 101070177.
\end{acks}

\bibliographystyle{ACM-Reference-Format}
\bibliography{sample-base}


\end{document}